\pdfoutput=1
\documentclass[11pt]{article}

\usepackage[final]{coling}

\usepackage{times}
\usepackage{latexsym}

\usepackage[T1]{fontenc}
\usepackage[utf8]{inputenc}

\usepackage{microtype}

\usepackage{inconsolata}

\usepackage{graphicx}

\usepackage{inconsolata}
\usepackage{inconsolata}
\usepackage{graphicx}
\usepackage{amsmath}
\usepackage{amssymb}
\usepackage{booktabs}
\usepackage{multirow}
\usepackage{subfigure}
\usepackage{booktabs}
\usepackage{pgfplots}
\usepackage{textcomp}
\usepackage{multirow}
\usepackage{mathtools}
\usepackage{amsthm}
\usepackage{csquotes}
\usepackage{subfigure}
\usepackage{xcolor}
\usepackage{enumitem}

\definecolor{mygreen}{rgb}{0.45,0.7,0.285}
\definecolor{myblue}{rgb}{0.18, 0.459, 0.71}

\title{Towards Understanding Multi-Task Learning (Generalization) of LLMs via Detecting and Exploring Task-Specific Neurons}

\author{
Yongqi Leng and Deyi Xiong {\thanks{~Corresponding author.}} \\
  College of Intelligence and Computing, Tianjin University, Tianjin, China\\
  \texttt{\{lengyq,dyxiong\}@tju.edu.cn} \\}

\begin{document}
\maketitle
\begin{abstract}
While large language models (LLMs) have demonstrated superior multi-task capabilities, understanding the learning mechanisms behind this is still a challenging problem. In this paper, we attempt to understand such mechanisms from the perspective of neurons. Specifically, we detect task-sensitive neurons in LLMs via gradient attribution on task-specific data. Through extensive deactivation and fine-tuning experiments, we demonstrate that the detected neurons are highly correlated with the given task, which we term as task-specific neurons. With these identified task-specific neurons, we delve into two common problems in multi-task learning and continuous learning: Generalization and Catastrophic Forgetting. We find that the overlap of task-specific neurons is strongly associated with generalization and specialization across tasks. Interestingly, at certain layers of LLMs, there is a high similarity in the parameters of different task-specific neurons, and such similarity is highly correlated with the generalization performance. Inspired by these findings, we propose a neuron-level continuous fine-tuning method that only fine-tunes the current task-specific neurons during continuous learning, and extensive experiments demonstrate the effectiveness of the proposed method. Our study provides insights into the interpretability of LLMs in multi-task learning.
\end{abstract}

\section{Introduction}
The advent and development of LLMs have marked a significant milestone in natural language processing \citep{brown2020language, touvron2023llama, achiam2023gpt}. LLMs perform instruction tuning on a wide range of tasks \citep{wei2021finetuned}, exhibiting superior capabilities across multiple tasks and even being able to generalize to unseen tasks \citep{sanh2021multitask}. Despite their effectiveness, the multi-task learning mechanisms of LLMs remain as an open question.

Previous studies have demonstrated the existence of language-related neurons in multilingual large language models (MLLMs), and these neurons have been analyzed to explore the multilingual learning mechanisms of MLLMs \citep{tang2024language, chen2024journey}. In contrast, research into the multi-task learning mechanisms of LLMs remains limited. We argue that multilingual learning is essentially a type of multi-task learning as well. Inspired by these studies and thinking analogously, we attempt to ask three questions: (1) Do task-related neurons exist in LLMs, from a broad perspective? (2) If they exist, can they facilitate the understanding of the multi-task learning mechanisms in LLMs? And (3) can we improve LLMs by exploring such neurons? 

In order to answer these questions, we perform neuronal analysis for LLMs. First, we identify neurons that are highly correlated with a given task by the gradient attribution method \citep{simonyan2013deep}. Subsequently, we conduct fine-tuning and deactivation experiments on these neurons, to analyze their impact on the performance of the given task. Results of extensive experiments show that task-related neurons are indeed present in LLMs and they are highly correlated with specific tasks. We hence term them as task-specific neurons.

With identified task-specific neurons, we delve into two problems in multi-task learning and continuous learning: Generalization and Catastrophic Forgetting. A well-developed deep learning system should have less forgetfulness about learned tasks, as well as a good ability to generalize to unseen tasks \citep{rish2021}. Therefore, we believe that analyzing these two problems in depth will contribute to enhance our further understanding of multi-task learning mechanisms in LLMs.

For this, we control the proportion of fine-tuned task-specific neurons to investigate generalization across tasks. We find that the overlap of task-specific neurons among different tasks is strongly correlated with generalization across these tasks, with higher overlap leading to higher generalization. However, in some cases, this overlap does not lead deterministically to generalization, since generalization is complex in nature, rather than a one-factor outcome. In addition to this, we find that at certain layers of LLMs, there is a high similarity between other task-specific neuron parameters and the task-specific neuron parameters of the task to be generalized, which suggests that LLMs learn to share knowledge between tasks, and that this similarity is highly correlated with the generalization results.

In the analysis of generalization, we not only observe the generalization across tasks, but also find that multi-task learning affects the performance of single-task specialization, which is caused by parameter interference between tasks. However, the cause of catastrophic forgetting is also parameter interference. Based on this, we propose a \textbf{N}euron-level \textbf{C}ontinuous \textbf{F}ine-\textbf{T}uning method (NCFT). Experimental results on two continuous learning benchmarks show that NCFT is capable of effectively mitigating catastrophic forgetting.

In summary, the main contributions of our study are as follows:
\begin{itemize}

\item We discover task-specific neurons in LLMs empirically through extensive experiments.
\item We provide significant insights into generalization across tasks with our task-specific neuron analysis.
\item We propose a neuron-level continuous learning fine-tuning method for mitigating catastrophic forgetting, and experiments demonstrate its effectiveness.
\end{itemize}

\section{Related Work}
\paragraph{Neuronal Interpretability}
With the development of LLMs, neuronal interpretability has gained much attention in recent years \citep{luo2024understanding, DBLP:journals/corr/abs-2309-15025}. Existing researches include knowledge storage \citep{dai2021knowledge}, knowledge conflicts mitigation \citep{DBLP:journals/corr/abs-2406-18406}, task solving \citep{wang2022finding}, sentiment analysis \citep{radford2017learning}, privacy preservation \citep{chen2024learnable, wu2023depn, DBLP:conf/acl/WuDXX24}, and model editing \citep{gu2023neuron}. In MLLMs, studies find the existence of language-related neurons and utilize neuronal analysis to reveal the multilingual mechanisms of MLLMs \citep{tang2024language, chen2024journey, zhao2024large}, which greatly contributes to the understanding of MLLMs. In contrast, limited studies are conducted on the neuronal analysis in multi-task learning in LLMs. We hence extend this line of research from multilingual learning to multi-task learning.

\paragraph{Cross-task Generalization}
\citet{wei2021finetuned} find that LLMs have excellent zero-shot performance after multi-task fine-tuning, which motivates further investigation into cross-task generalization in depth \citep{hupkes2022state, grosse2023studying}. Existing studies have shown that model size \citep{wei2021finetuned}, number of tasks \citep{sanh2021multitask}, and data quality \citep{zhou2024lima} all affect the performance of generalization, which illustrates that generalization is affected by a variety of factors. There are also some studies that aim to improving the generalization ability of LLMs, such as step-by-step instruction tuning \citep{wu2023improving} and hierarchical curriculum learning training strategy \citep{huang2024laying}. In addition to this, \citet{yang2024unveiling} conduct an empirical study to investigate generalization between tasks at a fine-grained level. Compared to the above studies, we focus more on the provenance of the generalization phenomenon after instruction tuning, and we analyze task-specific neurons to interpret generalization.

\paragraph{Catastrophic Forgetting}
Consistent with previous works \citep{ke2022continual, wang2024comprehensive}, we categorize continuous learning methods into three classes. (1) \textit{Rehearsal-based methods} mitigate forgetting by replaying data from previous tasks \citep{su2019generative}. (2) \textit{Regularization-based methods} add explicit regularization terms so that knowledge of previous tasks is retained during continuous training \citep{aljundi2018memory}. (3) \textit{Parameter isolation-based methods} assign task-specific parameters to new tasks, thereby reducing interference between tasks \citep{razdaibiedina2023progressive, wang2023rehearsal}. Our proposed NCFT method follows the philosophy of parameter isolation continuous learning, but unlike prior works, we do not need to introduce additional parameters and also consider the correlation between tasks.

\begin{figure*}[t]
\centering
% \raggedright
\includegraphics[width=1.0\linewidth]{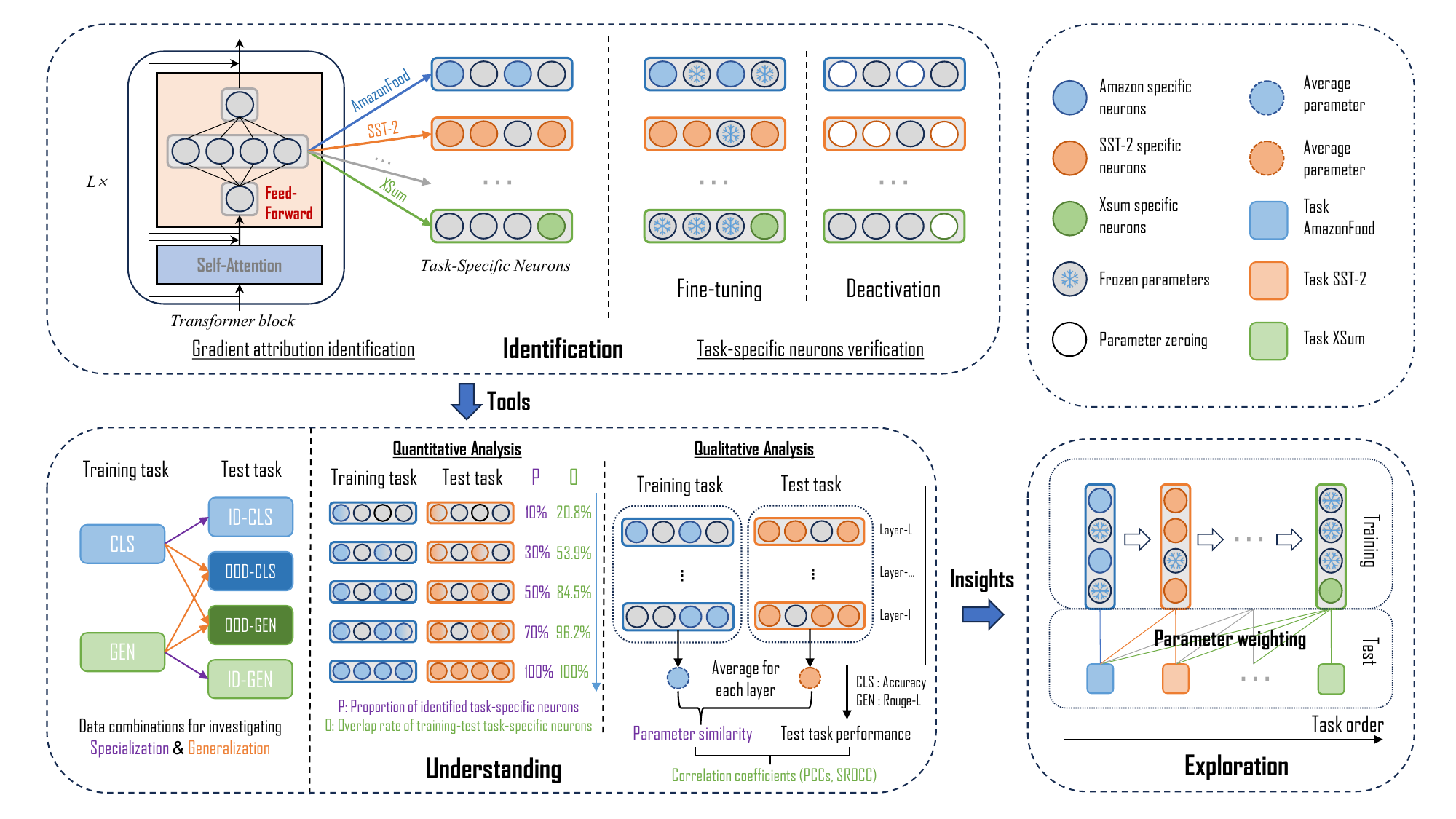}
\caption{Illustration of our research methodology. The entire framework consists of three components: Identification (task-specific neurons), Understanding (multi-task learning mechanisms of LLMs from the neuron level) and Exploration (neuron-level continuous fine-tuning method). The first component provides tools for mechanism understanding which in turn provides insights for the third component Exploration.}
\label{fig1}
\end{figure*}

\section{Methodology}
\label{sec:3}
Figure \ref{fig1} illustrates our proposed methodology. First, we compute task relevance scores for all neurons using the gradient attribution method. Based on these scores, we assign neurons to specific tasks to identify task-specific neurons. Next, we analyze these identified neurons both quantitatively and qualitatively to gain insights into the multi-task learning mechanisms of LLMs. Finally, capitalizing on our analysis of task-specific neurons, we propose a neuron-level continuous fine-tuning method designed to mitigate catastrophic forgetting in LLMs.

\subsection{Identifying Task-Specific Neurons in LLMs}
\label{sec:3.1}
To identify neurons highly relevant to a specific task, it is essential to determine the relevance of each neuron to task-specific data. Drawing inspiration from importance-based neuron fine-tuning studies \citep{xu2024let} and neuronal interpretability research \citep{tang2024language}, we employ the gradient attribution method to quantify each neuron's relevance score for a given task.

First, we need to clarify what we define as a neuron. Currently, the dominant architecture for LLMs is the auto-regressive transformer, in which the basic modules are multi-head self-attention (MHA) and feed-forward network (FFN). Here, we focus only on FFN, which have been demonstrated to store a large amount of parametric knowledge \citep{dai2021knowledge}.

The FFN module at layer $i$ can be formulated as:
\begin{equation}
\boldsymbol{h}^i=f(\boldsymbol{\tilde{h}}^i\boldsymbol{W}_1^i)\cdot \boldsymbol{W}_2^i
\label{equ:1}
\end{equation}
where $\boldsymbol{\tilde{h}}^i \in \mathbb{R}^d$ denotes the output of the MHA module in layer $i$, which is also the input of the current FFN module. $\boldsymbol{h}^i \in \mathbb{R}^d$ denotes the output of the current FFN module. $\boldsymbol{W}^i_1 \in \mathbb{R}^{d\times 4d}$ and $\boldsymbol{W}^i_2 \in \mathbb{R}^{4d\times d}$ are the parameters, and $f$ is the activation function.

A neuron is defined as a column in $\boldsymbol{W}^i_1$ or $\boldsymbol{W}^i_2$. Subsequently, we define the relevance score $\mathcal{R}^i_j$ of the $j$-th neuron in the $i$-th layer to a certain task:
\begin{equation}
\mathcal{R}^i_j=|\Delta\mathcal{L}(\boldsymbol{\omega}^i_j)|
\label{equ:2}
\end{equation}
where $\boldsymbol{\omega}^i_j$ is the output of the $j$-th neuron in the $i$-th layer, and $\Delta\mathcal{L}(\boldsymbol{\omega}^i_j)$ is the change in loss between setting $\boldsymbol{\omega}^i_j$ to 0 and keeping its original value. It can be converted to the following form by Taylor Expansion (see Appendix \ref{sec:appendix A-111} for detailed proof):
\begin{equation}
\mathcal{R}^i_j=\left|\Delta\mathcal{L}(\boldsymbol{\omega}^i_j)\right|=\left|\frac{\partial{\mathcal{L}}}{\partial{\boldsymbol{\omega}^i_j}} \boldsymbol{\omega}^i_j\right|
\label{equ:3}
\end{equation}

Subsequently, we take the neurons with the top $k\%$ relevance scores for the current task as task-specific neurons, where $k$ is a predefined hyper-parameter.

\subsection{Understanding Multi-Task Learning in LLMs by Analyzing Task-Specific Neurons}
\label{sec:3.2}

\begin{table*}[t]
\centering
\resizebox{0.9\textwidth}{!}{
\begin{tabular}{l|cccccc|c}
   \toprule
   \textbf{Method \textbackslash \ Task-CLS} & \textbf{AmazonFood} & \textbf{SST-2} & \textbf{QQP} &  \textbf{Paws} & \textbf{MNLI} & \textbf{GPTNLI} & \textbf{Avg.}\\
   \midrule
   Original & 91.8 & 92.4 & 83.2 & 91.6 & 84.8 & 82.4 & 87.7 \\
   Deactivate-Random & 90.6 & 91.2 & 79.8 & 87.6 & 80.5 & 79.3 & 84.8 \\
   Deactivate-Task & \textbf{83.6} & \textbf{84.6} & \textbf{72.8} & \textbf{70.2} & \textbf{73.3} & \textbf{71.4} & \textbf{76.0}  \\
   \toprule
   \textbf{Method \textbackslash \ Task-GEN} & \textbf{Sciqa} & \textbf{Tweetqa} & \textbf{E2E} &  \textbf{CommonGen} & \textbf{CNN/DailyMail} & \textbf{XSum} & \textbf{Avg.}\\
   \midrule
   Original & 54.3 & 45.6 & 52.6 & 49.8 & 34.7 & 36.8 & 45.6   \\
   Deactivate-Random & 50.8 & 41.3 & 48.7 & 47.3 & 31.3 & 34.4 & 42.3  \\
   Deactivate-Task & \textbf{33.6} & \textbf{29.3} & \textbf{39.6} & \textbf{37.8} & \textbf{25.5} & \textbf{26.3} & \textbf{32.0}  \\
   \bottomrule
\end{tabular}}
\caption{Performance of Llama-2-7b after task-specific neurons deactivation or without deactivation in each task. \enquote{Original} is the performance after fine-tuning with multi-task data without any neurons being deactivated. \enquote{Deactivate-Task} indicates deactivation of task-specific neurons. \enquote{Deactivate-Random} indicates that the same number of neurons are randomly selected for deactivation. Task-CLS: Classification Task. Task-GEN: Generation Task.}
\label{tab_1}
\end{table*}

\begin{table*}[t]
\centering
\resizebox{0.9\textwidth}{!}{
\begin{tabular}{l|cccccc|c}
   \toprule
   \textbf{Method \textbackslash \ Task-CLS} & \textbf{AmazonFood} & \textbf{SST-2} & \textbf{QQP} &  \textbf{Paws} & \textbf{MNLI} & \textbf{GPTNLI} & \textbf{Avg.}\\
   \midrule
   Zero-shot & 85.2 & 78.3 & 42.1 & 46.5 & 35.3 & 32.4 & 53.3 \\
   Train-Random & 85.5 & 80.3 & 45.6 & 47.8 & 34.7 & 34.8 & 54.8 \\
   Train-Task & \textbf{88.5} & \textbf{87.8} & \textbf{79.2} & \textbf{84.8} & \textbf{82.5} & \textbf{76.3} & \textbf{83.2}  \\
   \toprule
   \textbf{Method \textbackslash \ Task-GEN} & \textbf{Sciqa} & \textbf{Tweetqa} & \textbf{E2E} &  \textbf{CommonGen} & \textbf{CNN/DailyMail} & \textbf{XSum} & \textbf{Avg.}\\
   \midrule
   Zero-shot & 21.3 & 6.9 & 36.5 & 26.8 & 14.7 & 12.3 & 19.8   \\
   Train-Random & 22.8 & 11.8 & 37.4 & 29.6 & 17.7 & 15.8 & 22.5  \\
   Train-Task & \textbf{45.3} & \textbf{37.1} & \textbf{42.7} & \textbf{36.8} & \textbf{29.8} & \textbf{30.3} & \textbf{37.0}  \\
   \bottomrule
\end{tabular}}
\caption{Performance of Llama-2-7b after fine-tuning task-specific neurons and under the zero-shot setting. \enquote{Train-Task} indicates training task-specific neurons. \enquote{Train-Random} indicates that the same number of neurons are randomly selected for training. Task-CLS: Classification Task. Task-GEN: Generation Task.}
\label{tab_2}
\end{table*}

Once the presence of task-specific neurons is established, we proceed to analyze these neurons to understand the multi-task learning mechanisms of LLMs. First, we fine-tune varying proportions of task-specific neurons to study the impact on cross-task generalization and single-task specialization, exploring multi-task learning from a quantitative perspective. Additionally, we analyze the similarity between task-specific neuron parameters across different tasks, which encapsulate task-specific knowledge. In doing so, we aim to understand the provenance of generalization, thus revealing the multi-task learning mechanisms from a qualitative perspective.

In quantitative analysis, we set up different neuron proportions to investigate the trends in specialization and generalization. During fine-tuning, only the neurons specific to the current training task are trained. We use the results on the test set of the training task (in-domain, ID) to denote specialization performance, while the results on the test sets of other tasks (out-of-domain, OOD) to denote generalization performance.

In qualitative analysis, we compute the task-specific neuron parameters cosine similarity within a model between the task used to train that model and test task, and we study how this similarity varies across different layers of the model, aiming to investigate knowledge transfer between the test task and training task. In addition to this, we also compute the correlation coefficient between this parameter similarity and the performance on the corresponding test set, aiming to further demonstrate the association between parameter similarity and generalization.

\subsection{Exploring Task-Specific Neurons to Mitigate Catastrophic Forgetting of LLMs}
\label{sec:3.3}
Through the analysis of neurons, we find that while multi-task learning can effectively handle multiple tasks, it does not necessarily achieve optimal performance on a single task (see Section \ref{sec:5.1}). This is due to parameter interference between tasks. Similarly, catastrophic forgetting is also caused by parameter interference between tasks \citep{zhu2024landermt, wang2024comprehensive, wang2023rehearsal}. Inspired by this correlation, we propose that isolating task-specific neuron parameters during continuous training might mitigate catastrophic forgetting. In order to substantiate this, we introduce a neuron-level continuous fine-tuning method aimed at mitigating catastrophic forgetting in continuous learning.

Given a sequence of tasks ${D_1, \cdots, D_N}$, the tasks arrive sequentially in the order of the task sequence during the training stage. For the current task $D_n$, we update only the neuron-specific parameters of the current task, while keeping the other parameters frozen. During the test stage, the inference is executed as usual. We refer to this approach as \textbf{N}euron-level \textbf{C}ontinuous \textbf{F}ine-\textbf{T}uning (NCFT). This method isolates parameters for different tasks during training but maintains the original inference process. To better utilize the task-specific parameters of the already trained tasks, we propose using task similarity to weight different task-specific neurons during inference. We refer to this approach as \textbf{W}eighted \textbf{N}euron-level \textbf{C}ontinuous \textbf{F}ine-\textbf{T}uning (W-NCFT), more details of which are provided in Appendix \ref{sec:appendix A-222}.

\section{Experiments: Identifying Task-Specific Neurons}
\label{sec:4}
In this section, we conducted two groups of experiments to examine the existence of task-specific neurons as defined in Section \ref{sec:3.1}. 

\subsection{Experimental Setup}
\label{sec:4.1}
In the first group of experiments, we deactivated task-specific neurons to conduct deactivation experiments. Specifically, the deactivation was achieved by setting the activation value of these neurons to zero or by directly setting the corresponding parameter to zero. In the second group of experiments, we fine-tuned the task-specific neurons to carry out fine-tuning experiments. Particularly, only task-specific neurons were updated with parameters and other neurons were frozen during training. For both groups of experiments, we set the hyper-parameter $k=10$.

We tested two open-source models that perform well on multi-tasks, including Llama-2-7b \citep{touvron2023llama} and Bloom-7b1 \citep{le2023bloom}. We tested two main types of tasks: \textit{classification} and \textit{generation}, details of the dataset and evaluation metrics can be
found in Appendix \ref{sec:appendix B-000}.

\subsection{Results}
\label{sec:4.1}
Table \ref{tab_1} shows the results of the deactivation experiments. Despite deactivating only $10\%$ task-specific neurons, it has a large negative impact on task-specific processing capacity. In contrast, deactivating the same number of randomly selected neurons resulted in a small impact.

To bolster the dependability of task-specific neurons, we conducted additional fine-tuning experiments. As shown in Table \ref{tab_2}, the fine-tuning approach to task-specific neurons yields remarkable improvements compared to the approach of fine-tuning randomly selected neurons ($29.9$ vs $1.5$ in classification tasks while $17.2$ vs $2.7$ in generation tasks). These improvements remain consistent across both task categories (classificatiton and generation). The only task where the improvement is not significant is AmazonFood, since it has a good enough zero-shot result. Appendix \ref{sec:appendix B-111} presents results for Bloom-7b1, which demonstrate the same trend. Additionally, we show the impact of inactivating or fine-tuning a particular class of task-specific neurons on other tasks in Appendix \ref{sec:appendix B-111}.

In summary, we find that the effects of fine-tuning and perturbing task-specific neurons are more significant than those of randomly selected neurons.  Consequently, we can empirically assert the presence of task-specific neurons within LLMs.

\begin{table*}[t]
\centering
\resizebox{0.7\textwidth}{!}{
\begin{tabular}{cccc}
   \toprule
   \textbf{Group} & \textbf{Training Tasks} & \textbf{ID Test Tasks} & \textbf{OOD Test Tasks} \\
   \midrule
   \multirow{2}{*}{(a)} & \multirow{2}{*}{Amazon, QQP, MNLI} & \multirow{2}{*}{\textcolor{myblue}{Amazon, QQP, MNLI}} & \textcolor{orange}{SST-2, Paws, GPTNLI}  \\
   ~ & ~ & ~ & \textcolor{mygreen}{Tweetqa, CommonGen, Xsum} \\
   \midrule
   \multirow{2}{*}{(b)} & \multirow{2}{*}{Sciqa, E2E, CNN} & \multirow{2}{*}{\textcolor{violet}{Sciqa, E2E, CNN}} & \textcolor{orange}{SST-2, Paws, GPTNLI} \\
   ~ & ~ & ~ & \textcolor{mygreen}{Tweetqa, CommonGen, Xsum} \\
   \bottomrule
\end{tabular}}
\caption{Experimental groups for exploring generalization and specialization. Results from the in-domain (ID) test set indicate generalization performance while results from the out-of-domain (OOD) test set indicate specialization performance. Four test set colors, corresponding to the legend in Figure \ref{fig2}. \textcolor{myblue}{Amazon, QQP, MNLI} corresponds to \textbf{ID-CLS} in the legend. \textcolor{violet}{Sciqa, E2E, CNN} corresponds to \textbf{ID-GEN} in the legend. \textcolor{orange}{SST-2, Paws, GPTNLI} corresponds to \textbf{OOD-CLS} in the legend. \textcolor{mygreen}{Tweetqa, CommonGen, Xsum} corresponds to \textbf{OOD-GEN} in the legend.}
\label{tab_3}
\end{table*}

\begin{figure*}[t]
\centering
\includegraphics[width=0.8\linewidth]{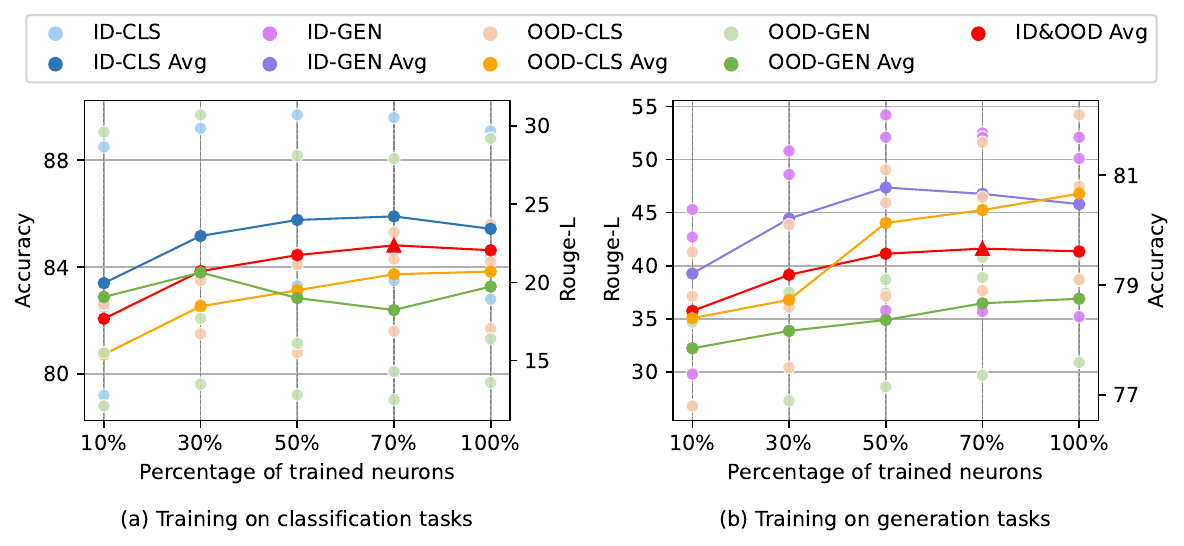}
\caption{Results on classification and generation tasks after fine-tuning different proportions of task-specific neurons. The red line indicates the average of the results on the in-domain (ID) test set and out-of-domain (OOD) test set with the same type of training task. For example, in subfigure (a), the red line shows the average of the blue and orange lines while the average of the purple and green lines in subfigure (b). The correspondence of the other colored lines to the test set is shown in the caption of Table \ref{tab_3}.}
\label{fig2}
\vspace{-1.0em}
\end{figure*}

\section{Experiments: Analyzing Task-Specific Neurons to Interpret Generalization}
\label{sec:5}
We analyzed task-specific neurons to understand the multi-task learning mechanisms of LLMs. Based on the analytical approach of Section \ref{sec:3.2}, we conducted two sets of experiments, qualitative and quantitative, on various training-test combinations listed in Table \ref{tab_3}.

\subsection{Proportion of Task-Specific Neurons}
\label{sec:5.1}
We controlled the proportion of fine-tuned task-specific neurons to conduct experiments on the various training-test combinations. Figure \ref{fig2} shows results for all training-test combinations. In each subfigure, we focus only on the trend of each color line. Comparisons between different color lines are meaningless because they represent different tasks.

\paragraph{Specialization.}
As the proportion of trained task-specific neurons increases, the specialization performance (see Section \ref{sec:3.2} for definition) for both classification and generation tasks first ascends and then declines, reaching its peak at $70\%$ for the classification task (blue line in Figure \ref{fig2} (a)) and at $50\%$ for the generation task (purple line in Figure \ref{fig2} (b)). This is contrary to our intuition that under normal circumstances, better results should be obtained as more task-specific neurons are trained. We analyzed the reason behind this, which could stem from the parameter interference between different tasks induced by simultaneous training of three tasks. This interference further results in the specialization performance of a single task not exhibiting a continuous improvement as more parameters are trained. To corroborate this, we conducted ablation experiments. Specifically, we trained a model for each task, meaning that the fine-tuning of task-specific neurons was conducted individually. Results are shown in Appendix \ref{sec:appendix B-222}, wherein we observe a continuous enhancement in performance as the proportion of neurons increases, thus validating our analysis.

\begin{figure*}[t]
\centering
\includegraphics[width=1.0\linewidth]{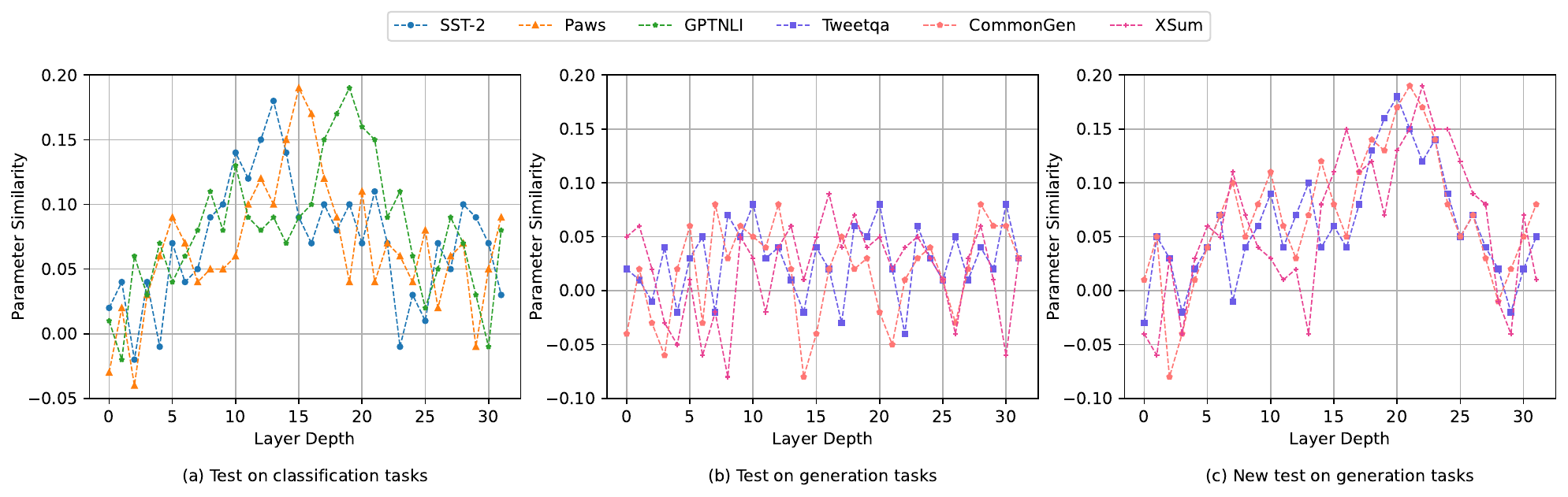}
\caption{The similarity of the task-specific neuron parameters between the test task and training tasks in different layers.}
\label{fig3}
\vspace{-1.0em}
\end{figure*}

\paragraph{Generalization.}
As the proportion of trained task-specific neurons increases, we find a continuous increasing trend for the performance of generalization from the trained classification tasks to other classification tasks (orange line in Figure \ref{fig2} (a)). Similarly, the performance of generalization from the trained generation tasks to other classification tasks (orange line in Figure \ref{fig2} (b)) and from the trained generation tasks to other generation tasks (green line in Figure \ref{fig2} (b)) shows the same trend. The overlap rate of task-specific neurons between the training and test tasks can be found in Appendix \ref{sec:appendix B-333}, where it becomes evident that as the proportion of trained task-specific neurons increases, the overlap rate also experiences a significant surge. Consequently, one plausible explanation is that the overlap of task-specific neurons contributes to transfer learning between tasks, ultimately resulting in consistently higher generalization performance. To this end, we conducted ablation experiments in Appendix \ref{sec:appendix B-444} to exclude the effect of the variable of the number of trained parameters, and the results support this conclusion. However, no generalization is produced from the trained classification tasks to other generation tasks (green line in Figure \ref{fig2} (a)), and the test results are similar to the zero-shot results in Table \ref{tab_2}. The reason for no generalization observed from classification to generation might be that classification tasks are usually easier than generation tasks as they only need to predict a single label. In contrast, generation tasks need to generate consecutive texts that satisfy the task requirements, which is relatively harder. This observation is consistent with that observed by \citet{yang2024unveiling}.

In summary, our findings reveal that when training all parameters of the model under the multi-task learning setup, inevitable interference among tasks occurs, thereby diminishing the efficacy of individual tasks to some degree. Furthermore, our experiments illustrate the efficacy of controlling the appropriate proportion of fine-tuned task-specific neurons as a promising strategy. Additionally, we observe a significant correlation between the overlap of task-specific neurons and generalization performance across tasks. However, this overlap does not always guarantee deterministic generalization, as numerous factors also play pivotal roles. These comprehensive analyses serve to enrich our comprehension on generalization.

\subsection{Parameters of Task-Specific Neurons}
\label{sec:5.2}
We evaluated the similarity of specific neuron parameters for the training and test tasks (see Section \ref{sec:3.2} for the way to calculate the similarity) aiming to conduct a qualitative analysis of generalization provenance. We trained a separate model (full-parameter training) for each of the six training tasks in the training-test combination in Table \ref{tab_3}, denoted as $M_1, \cdots , M_6$. We then tested these models on the six out-of-domain test tasks listed in that combination, denoted as $T_1, \cdots , T_6$. In a particular layer, for model $M_i$ and test task $T_j$, $\boldsymbol{P}^i_i$ and $\boldsymbol{P}^i_j$ are used to denote the task-specific neuron parameters of training task $i$ and test task $j$ in $M_i$, respectively. Then, we calculated the cosine similarity between $\boldsymbol{P}^i_i$ and $\boldsymbol{P}^i_j$. For test task $T_j$, testing across the six trained models provides six similarity measures. We computed the average of these similarities and then investigated how this average similarity varies across different layers of the model, aiming to show knowledge transfer to the test task $T_j$. Figure \ref{fig3} illustrates the similarity of the different layers for three different settings.

\begin{table*}[t]
\centering
\resizebox{0.9\textwidth}{!}{
\begin{tabular}{lcccccccccccc}
   \toprule
   \multirow{2}{*}{\textbf{Testset}} & \multicolumn{2}{c}{\textbf{SST-2}} &  \multicolumn{2}{c}{\textbf{Paws}} &  \multicolumn{2}{c}{\textbf{GPTNLI}} &   \multicolumn{2}{c}{\textbf{Tweetqa}} &  \multicolumn{2}{c}{\textbf{CommonGen}} &  \multicolumn{2}{c}{\textbf{Xsum}}\\
   \cmidrule(r){2-3} \cmidrule(r){4-5} \cmidrule(r){6-7} \cmidrule(r){8-9} \cmidrule(r){10-11} \cmidrule(r){12-13}
   ~ & \textbf{r} & \textbf{p-value} & \textbf{r} & \textbf{p-value} & \textbf{r} & \textbf{p-value} & \textbf{r} & \textbf{p-value} & \textbf{r} & \textbf{p-value} & \textbf{r} & \textbf{p-value} \\
   \midrule
   PCCs & 0.87 & 0.02 & 0.92 & 0.01 & 0.79 & 0.05 & 0.96 & 0.00 & 0.96 & 0.00 & 0.97 & 0.00\\
   SROCC & 0.81 & 0.05 & 0.77 & 0.07 & 0.81 & 0.05 & 0.77 & 0.07 & 0.83 & 0.04 & 0.71 & 0.11 \\
   \bottomrule
\end{tabular}}
\caption{Correlation coefficients between the similarity of specific neuron parameters and generalization performance. PCCs denotes Pearson correlation coefficients and SROCC denotes Spearman correlation coefficients.}
\label{tab_4}
\end{table*}

\paragraph{Parameter Similarity on Classification Tasks.}
Figure \ref{fig3} (a) shows how the parameter similarity across three classification test tasks. We find that at the bottom layer, the similarity remains notably low. When reaching a certain layer depth, similarity starts to gradually increase. Finally, the similarity drops again to the value close to that at the bottom layer. This observation holds for all three classification tasks. This illustrates that a model learns the shared knowledge between tasks only after a certain number of layers. In this aspect, knowledge transfer occurs, thus contributing to generalization. \citet{chatterjee2024language} provide similar findings in cross-task in-context learning to ours, which show that information transfer across tasks occurs only after a certain layer depth is reached. Although their findings are based on in-context learning, in-context learning can be understood as a form of implicit training without parameter updates \citep{akyurek2022learning, von2023transformers}. We consider these findings resonate with each other.

\paragraph{Parameter Similarity on Generation Tasks.}
However, on the three generation test tasks in Figure \ref{fig3} (b), we find no such trend. In Section \ref{sec:5.1}, we have previously found that it is difficult to generalize from classification tasks to generation tasks. Therefore, we conjecture that the absence of the expected observation in Figure \ref{fig3} (b) is due to the fact that the six training models used include three models trained with classification tasks, which do not have good parameter similarity within these three models. In turn, after averaging the parameter similarity, lower values appear. To substantiate this conjecture, we tested again using three of the six models trained with generation tasks. Results are shown in Figure \ref{fig3} (c), and the overall trend is similar to that observed in Figure \ref{fig3} (a). Only the layer depths where the similarity rises differ, which indicates that the location where knowledge transfer occurs varies across tasks. At the same time, this confirms our conjecture.

\paragraph{Parameter Similarity and Generalization.}
We further investigated the relationship between the similarity of task-specific neuron parameters and generalization performance. For each test task, we used six models. We then calculated the similarity in each model between the specific neuron parameters of that test task and the specific neuron parameters of the training task used by that model. Finally, we calculated the correlation coefficients between these parameter similarities and the predictions of the six models. As shown in Table \ref{tab_4}, we find that the similarity is highly correlated with the generalization performance.

In summary, our findings suggest a correlation between the generalization across different tasks and the similarity of task-specific neuron parameters.  When layers after a certain depth are reached, the model can learn shared knowledge between tasks, which contributes to the generalization across these tasks. Additionally, higher parameter similarity corresponds to better generalization performance. Our conclusions provide a guideline for improving generalization performance across tasks.

\section{Experiments: Fine-tuning Task-specific Neurons to Mitigate Catastrophic Forgetting}
\label{sec:6}

We finally conducted experiments on the two benchmarks of continuous learning so as to test the effectiveness of the NCFT and W-NCFT methods described in Section \ref{sec:3.3}.

\subsection{Experimental Setup}
\label{sec:6.1}
\paragraph{Model and Datasets}
We used Llama-2-7b as the model for experiments. We used two continuous learning benchmarks, Standard CL Benchmark and Large Number of Tasks Benchmark \citep{razdaibiedina2023progressive}, and tested different task orders. Details on the datasets and task order can be found in Appendix \ref{sec:appendix C-111}.

\paragraph{Metrics}
We used continuous learning performance and forgetting rate as evaluation metrics. Let $a_{i, j}$ be the testing accuracy of the $i$-th task after training on $j$-th task, and $A_i$ denote the testing accuracy after training on task $i$ alone. The evaluation metrics are:

\begin{table*}[t]
\centering
\resizebox{0.8\textwidth}{!}{
\begin{tabular}{l|cccc|cccc}
   \toprule
   \textbf{Method} & \textbf{Order-1} & \textbf{Order-2} & \textbf{Order-3} &  \textbf{Avg.} & \textbf{Order-4} & \textbf{Order-5} & \textbf{Order-6} & \textbf{Avg.}\\
   \midrule
   SeqFT & 46.4 & 47.3 & 47.5 & 47.1 & 35.6 & 34.8 & 33.5 & 34.6 \\
   SeqLoRA & 53.6 & 54.8 & 53.1 & 53.8 & 47.9 & 49.5 & 45.7 & 47.7 \\
   EPI & 48.1 & 48.0 & 49.0 & 48.4 & 42.3 & 41.8 & 43.6 & 42.6 \\
   O-LoRA & \textbf{76.8} & \textbf{75.7} & \textbf{75.7} & \textbf{76.1} & \textbf{73.7} & 69.2 & 72.0 & 71.6 \\
   NCFT (Ours) & 71.3 & 70.9 & 71.6 & 71.3 & 70.5 & 68.3 & 71.2 & 70.0  \\
   W-NCFT (Ours) & 73.7 & 72.3 & 73.8 & 73.3 & 73.4 & \textbf{70.1} & \textbf{72.6} & \textbf{72.0} \\
   \midrule
   Per-Task FT & 77.2 & 77.2 & 77.2 & 77.2 & 84.5 & 84.5 & 84.5 & 84.5 \\
   \bottomrule
\end{tabular}}
\caption{Results on two continual learning benchmarks. The average accuracy after training on the last task is reported.}
\label{tab_5}
\vspace{-1.0em}
\end{table*}

\begin{itemize}
\item \textbf{Performance on Continuous Learning (CL).} 
The average accuracy of all tasks after training on the last task, is computed as:
\begin{equation}
\mathrm{CL}=\frac{1}{N} \sum_{i=1}^{N}a_{i, N}
\label{equ:12}
\end{equation}
\item \textbf{Forgetting (FG).}
Following the evaluation metrics proposed by \citet{scialom2022fine}, we utilized relative gain to calculate the forgetting rate at different stages. The forgetting rate for the $j$-th stage is calculated as:
\begin{equation}
\mathrm{FG}_j=\frac{1}{j-1}\sum_{i=1}^{j-1}\frac{a_{i, j}}{A_i} \times 100\%
\label{equ:13}
\end{equation}
\end{itemize}

\paragraph{Baselines}
We used the following continual learning techniques as baselines:
\begin{itemize}
\item \textbf{SeqFT}: training the entire model parameters on a sequence of tasks.
\item \textbf{SeqLoRA}: training fixed-size LoRA parameters on a sequence of tasks.
\item \textbf{EPI} \citep{wang2023rehearsal}: allocating a small portion of private parameters and learns them with a shared pre-trained model.
\item \textbf{O-LoRA} \citep{wang2023orthogonal}: learning tasks in different (low-rank) vector subspaces that are kept orthogonal to each other in order to minimize interference.
\item \textbf{Per-Task FT}: training a separate model for each task.
\end{itemize}

\subsection{Results and Analysis}
\label{sec:6.2}
\begin{figure}[t]
\centering
\includegraphics[width=0.9\linewidth]{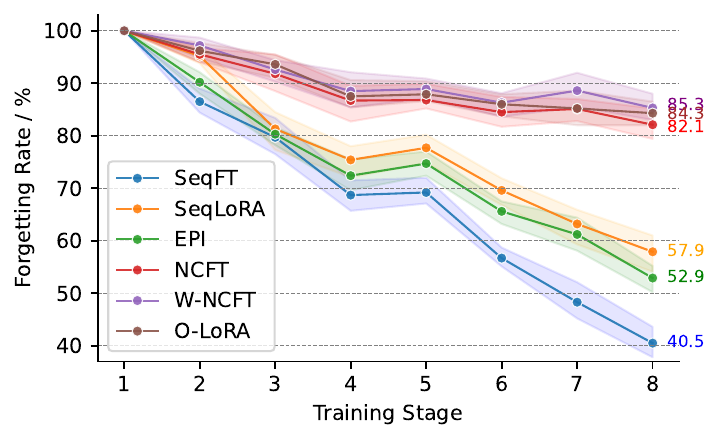}
\caption{Forgetting rates for eight stages on the Large Number of Tasks benchmark.}
\label{fig4}
\end{figure}

As shown in Table \ref{tab_5}, our proposed method achieves significant improvements compared to the first three baselines across both task benchmarks. Even comparing O-LoRA \citep{wang2023orthogonal} there is a very small difference, on the first benchmark, our method is inferior to O-LoRA, but we outperform it on the second benchmark. Such improvements are consistent across various sequences of tasks, illustrating the effectiveness and robustness of our approach. Additionally, we find that W-NCFT outperforms NCFT, suggesting that weighting different task-specific parameters based on their similarity enhances the performance of continuous learning. Figure \ref{fig4} illustrates the forgetting rate across eight stages on the Large Number of Tasks benchmark, and we can find that both NCFT and W-NCFT methods substantially mitigate catastrophic forgetting.

It is worth noting that although our proposed method effectively mitigates catastrophic forgetting, it still has some shortcomings. As shown in Figure \ref{fig4}, there remains a gap between the performance of the NCFT and W-NCFT methods and that of Per-Task FT. This indicates that catastrophic forgetting has not been entirely resolved. Additionally, W-NCFT employs task similarity to weight the parameters, which is a static approach. A dynamic weighting method, applied during continuous training, could potentially yield better results. Nevertheless, it is undeniable that this empirical study demonstrates the effectiveness of the task-specific parameter isolation approach in mitigating catastrophic forgetting.

\section{Conclusion}
In this study, we have presented a methodology framework for understanding multi-task learning and cross-task generalization of LLMs from the perspective of neurons. With this framework, we have conducted an extensive analysis of LLMs to identify task-specific neurons that are highly correlated with specific tasks. Using these task-specific neurons, we have investigated two common problems of LLMs in multi-task learning and continuous learning: generalization and catastrophic forgetting. Our findings indicate that the overlap of task-specific neurons is strongly associated with generalization. Furthermore, we find that the parameter similarity of these neurons reflects the degree of knowledge sharing, contributing to generalization. Additionally, we propose a neuron-level continuous fine-tuning method to effectively mitigate catastrophic forgetting. The proposed method only fine-tuning the current task-specific neurons in continuous learning, and experimental results in two continuous learning benchmarks demonstrate the effectiveness of our method.

\section*{Limitations}
Our analysis is based on the identification of neurons. In the identification experiments, we did not conduct a detailed analysis on the hyper-parameters, but only used empirical values. However, we believe that it is crucial to identify neurons more accurately, as this may better utilize neurons for these specific tasks. Additionally, our analysis of generalization is currently conducted on only classification and generation tasks. There is a need to extend this analysis to a broader range of tasks. We plan to address these more detailed studies in our future work.

\section*{Ethics Statement}
This study adheres to the ethical guidelines set forth by our institution and follows the principles outlined in the ACM Code of Ethics and Professional Conduct. All datasets used in our experiments are publicly available.

\section*{Acknowledgments}
The present research was supported by the National Key Research and Development Program of China (Grant No. 2023YFE0116400). We would like to thank the anonymous reviewers for their insightful comments.

% \section*{Acknowledgments}

% This document has been adapted
% by Steven Bethard, Ryan Cotterell and Rui Yan
% from the instructions for earlier ACL and NAACL proceedings, including those for
% ACL 2019 by Douwe Kiela and Ivan Vuli\'{c},
% NAACL 2019 by Stephanie Lukin and Alla Roskovskaya,
% ACL 2018 by Shay Cohen, Kevin Gimpel, and Wei Lu,
% NAACL 2018 by Margaret Mitchell and Stephanie Lukin,
% Bib\TeX{} suggestions for (NA)ACL 2017/2018 from Jason Eisner,
% ACL 2017 by Dan Gildea and Min-Yen Kan,
% NAACL 2017 by Margaret Mitchell,
% ACL 2012 by Maggie Li and Michael White,
% ACL 2010 by Jing-Shin Chang and Philipp Koehn,
% ACL 2008 by Johanna D. Moore, Simone Teufel, James Allan, and Sadaoki Furui,
% ACL 2005 by Hwee Tou Ng and Kemal Oflazer,
% ACL 2002 by Eugene Charniak and Dekang Lin,
% and earlier ACL and EACL formats written by several people, including
% John Chen, Henry S. Thompson and Donald Walker.
% Additional elements were taken from the formatting instructions of the \emph{International Joint Conference on Artificial Intelligence} and the \emph{Conference on Computer Vision and Pattern Recognition}.

\bibliography{custom}

\begin{thebibliography}{52}
\providecommand{\natexlab}[1]{#1}

\bibitem[{Aky{\"{u}}rek et~al.(2023)Aky{\"{u}}rek, Schuurmans, Andreas, Ma, and Zhou}]{akyurek2022learning}
Ekin Aky{\"{u}}rek, Dale Schuurmans, Jacob Andreas, Tengyu Ma, and Denny Zhou. 2023.
\newblock \href {https://openreview.net/forum?id=0g0X4H8yN4I} {What learning algorithm is in-context learning? investigations with linear models}.
\newblock In \emph{The Eleventh International Conference on Learning Representations, {ICLR} 2023, Kigali, Rwanda, May 1-5, 2023}. OpenReview.net.

\bibitem[{Aljundi et~al.(2018)Aljundi, Babiloni, Elhoseiny, Rohrbach, and Tuytelaars}]{aljundi2018memory}
Rahaf Aljundi, Francesca Babiloni, Mohamed Elhoseiny, Marcus Rohrbach, and Tinne Tuytelaars. 2018.
\newblock \href {https://doi.org/10.1007/978-3-030-01219-9\_9} {Memory aware synapses: Learning what (not) to forget}.
\newblock In \emph{Computer Vision - {ECCV} 2018 - 15th European Conference, Munich, Germany, September 8-14, 2018, Proceedings, Part {III}}, volume 11207 of \emph{Lecture Notes in Computer Science}, pages 144--161. Springer.

\bibitem[{Brown et~al.(2020)Brown, Mann, Ryder, Subbiah, Kaplan, Dhariwal, Neelakantan, Shyam, Sastry, Askell, Agarwal, Herbert{-}Voss, Krueger, Henighan, Child, Ramesh, Ziegler, Wu, Winter, Hesse, Chen, Sigler, Litwin, Gray, Chess, Clark, Berner, McCandlish, Radford, Sutskever, and Amodei}]{brown2020language}
Tom~B. Brown, Benjamin Mann, Nick Ryder, Melanie Subbiah, Jared Kaplan, Prafulla Dhariwal, Arvind Neelakantan, Pranav Shyam, Girish Sastry, Amanda Askell, Sandhini Agarwal, Ariel Herbert{-}Voss, Gretchen Krueger, Tom Henighan, Rewon Child, Aditya Ramesh, Daniel~M. Ziegler, Jeffrey Wu, Clemens Winter, Christopher Hesse, Mark Chen, Eric Sigler, Mateusz Litwin, Scott Gray, Benjamin Chess, Jack Clark, Christopher Berner, Sam McCandlish, Alec Radford, Ilya Sutskever, and Dario Amodei. 2020.
\newblock \href {https://proceedings.neurips.cc/paper/2020/hash/1457c0d6bfcb4967418bfb8ac142f64a-Abstract.html} {Language models are few-shot learners}.
\newblock In \emph{Advances in Neural Information Processing Systems 33: Annual Conference on Neural Information Processing Systems 2020, NeurIPS 2020, December 6-12, 2020, virtual}.

\bibitem[{Chatterjee et~al.(2024)Chatterjee, Tanwar, Dutta, and Chakraborty}]{chatterjee2024language}
Anwoy Chatterjee, Eshaan Tanwar, Subhabrata Dutta, and Tanmoy Chakraborty. 2024.
\newblock \href {https://doi.org/10.18653/V1/2024.ACL-LONG.621} {Language models can exploit cross-task in-context learning for data-scarce novel tasks}.
\newblock In \emph{Proceedings of the 62nd Annual Meeting of the Association for Computational Linguistics (Volume 1: Long Papers), {ACL} 2024, Bangkok, Thailand, August 11-16, 2024}, pages 11568--11587. Association for Computational Linguistics.

\bibitem[{Chen et~al.(2024{\natexlab{a}})Chen, Hu, Feng, and Liu}]{chen2024learnable}
Ruizhe Chen, Tianxiang Hu, Yang Feng, and Zuozhu Liu. 2024{\natexlab{a}}.
\newblock \href {https://aclanthology.org/2024.acl-short.25} {Learnable privacy neurons localization in language models}.
\newblock In \emph{Proceedings of the 62nd Annual Meeting of the Association for Computational Linguistics, {ACL} 2024 - Short Papers, Bangkok, Thailand, August 11-16, 2024}, pages 256--264. Association for Computational Linguistics.

\bibitem[{Chen et~al.(2024{\natexlab{b}})Chen, Cao, Chen, Liu, and Zhao}]{chen2024journey}
Yuheng Chen, Pengfei Cao, Yubo Chen, Kang Liu, and Jun Zhao. 2024{\natexlab{b}}.
\newblock \href {https://doi.org/10.1609/AAAI.V38I16.29735} {Journey to the center of the knowledge neurons: Discoveries of language-independent knowledge neurons and degenerate knowledge neurons}.
\newblock In \emph{Thirty-Eighth {AAAI} Conference on Artificial Intelligence, {AAAI} 2024, Thirty-Sixth Conference on Innovative Applications of Artificial Intelligence, {IAAI} 2024, Fourteenth Symposium on Educational Advances in Artificial Intelligence, {EAAI} 2014, February 20-27, 2024, Vancouver, Canada}, pages 17817--17825. {AAAI} Press.

\bibitem[{Dai et~al.(2022)Dai, Dong, Hao, Sui, Chang, and Wei}]{dai2021knowledge}
Damai Dai, Li~Dong, Yaru Hao, Zhifang Sui, Baobao Chang, and Furu Wei. 2022.
\newblock \href {https://doi.org/10.18653/V1/2022.ACL-LONG.581} {Knowledge neurons in pretrained transformers}.
\newblock In \emph{Proceedings of the 60th Annual Meeting of the Association for Computational Linguistics (Volume 1: Long Papers), {ACL} 2022, Dublin, Ireland, May 22-27, 2022}, pages 8493--8502. Association for Computational Linguistics.

\bibitem[{Dusek et~al.(2020)Dusek, Novikova, and Rieser}]{duvsek2020evaluating}
Ondrej Dusek, Jekaterina Novikova, and Verena Rieser. 2020.
\newblock \href {https://doi.org/10.1016/J.CSL.2019.06.009} {Evaluating the state-of-the-art of end-to-end natural language generation: The {E2E} {NLG} challenge}.
\newblock \emph{Comput. Speech Lang.}, 59:123--156.

\bibitem[{Grosse et~al.(2023)Grosse, Bae, Anil, Elhage, Tamkin, Tajdini, Steiner, Li, Durmus, Perez, Hubinger, Lukosiute, Nguyen, Joseph, McCandlish, Kaplan, and Bowman}]{grosse2023studying}
Roger~B. Grosse, Juhan Bae, Cem Anil, Nelson Elhage, Alex Tamkin, Amirhossein Tajdini, Benoit Steiner, Dustin Li, Esin Durmus, Ethan Perez, Evan Hubinger, Kamile Lukosiute, Karina Nguyen, Nicholas Joseph, Sam McCandlish, Jared Kaplan, and Samuel~R. Bowman. 2023.
\newblock \href {https://doi.org/10.48550/ARXIV.2308.03296} {Studying large language model generalization with influence functions}.
\newblock \emph{CoRR}, abs/2308.03296.

\bibitem[{Gu et~al.(2023)Gu, Chen, and Aleti}]{gu2023neuron}
Jian Gu, Chunyang Chen, and Aldeida Aleti. 2023.
\newblock \href {https://doi.org/10.48550/ARXIV.2312.05356} {Neuron patching: Neuron-level model editing on code generation and llms}.
\newblock \emph{CoRR}, abs/2312.05356.

\bibitem[{Hermann et~al.(2015)Hermann, Kocisk{\'{y}}, Grefenstette, Espeholt, Kay, Suleyman, and Blunsom}]{hermann2015teaching}
Karl~Moritz Hermann, Tom{\'{a}}s Kocisk{\'{y}}, Edward Grefenstette, Lasse Espeholt, Will Kay, Mustafa Suleyman, and Phil Blunsom. 2015.
\newblock \href {https://proceedings.neurips.cc/paper/2015/hash/afdec7005cc9f14302cd0474fd0f3c96-Abstract.html} {Teaching machines to read and comprehend}.
\newblock In \emph{Advances in Neural Information Processing Systems 28: Annual Conference on Neural Information Processing Systems 2015, December 7-12, 2015, Montreal, Quebec, Canada}, pages 1693--1701.

\bibitem[{Huang et~al.(2024)Huang, He, Xu, Liang, and Xiao}]{huang2024laying}
Yuncheng Huang, Qianyu He, Yipei Xu, Jiaqing Liang, and Yanghua Xiao. 2024.
\newblock \href {https://doi.org/10.48550/ARXIV.2403.09479} {Laying the foundation first? investigating the generalization from atomic skills to complex reasoning tasks}.
\newblock \emph{CoRR}, abs/2403.09479.

\bibitem[{Hupkes et~al.(2022)Hupkes, Giulianelli, Dankers, Artetxe, Elazar, Pimentel, Christodoulopoulos, Lasri, Saphra, Sinclair, Ulmer, Schottmann, Batsuren, Sun, Sinha, Khalatbari, Ryskina, Frieske, Cotterell, and Jin}]{hupkes2022state}
Dieuwke Hupkes, Mario Giulianelli, Verna Dankers, Mikel Artetxe, Yanai Elazar, Tiago Pimentel, Christos Christodoulopoulos, Karim Lasri, Naomi Saphra, Arabella Sinclair, Dennis Ulmer, Florian Schottmann, Khuyagbaatar Batsuren, Kaiser Sun, Koustuv Sinha, Leila Khalatbari, Maria Ryskina, Rita Frieske, Ryan Cotterell, and Zhijing Jin. 2022.
\newblock \href {https://doi.org/10.48550/ARXIV.2210.03050} {State-of-the-art generalisation research in {NLP:} a taxonomy and review}.
\newblock \emph{CoRR}, abs/2210.03050.

\bibitem[{Ke and Liu(2022)}]{ke2022continual}
Zixuan Ke and Bing Liu. 2022.
\newblock \href {https://doi.org/10.48550/ARXIV.2211.12701} {Continual learning of natural language processing tasks: {A} survey}.
\newblock \emph{CoRR}, abs/2211.12701.

\bibitem[{Keung et~al.(2020)Keung, Lu, Szarvas, and Smith}]{keung2020multilingual}
Phillip Keung, Yichao Lu, Gy{\"{o}}rgy Szarvas, and Noah~A. Smith. 2020.
\newblock \href {https://doi.org/10.18653/V1/2020.EMNLP-MAIN.369} {The multilingual amazon reviews corpus}.
\newblock In \emph{Proceedings of the 2020 Conference on Empirical Methods in Natural Language Processing, {EMNLP} 2020, Online, November 16-20, 2020}, pages 4563--4568. Association for Computational Linguistics.

\bibitem[{Lin et~al.(2020)Lin, Zhou, Shen, Zhou, Bhagavatula, Choi, and Ren}]{lin2019commongen}
Bill~Yuchen Lin, Wangchunshu Zhou, Ming Shen, Pei Zhou, Chandra Bhagavatula, Yejin Choi, and Xiang Ren. 2020.
\newblock \href {https://doi.org/10.18653/V1/2020.FINDINGS-EMNLP.165} {Commongen: {A} constrained text generation challenge for generative commonsense reasoning}.
\newblock In \emph{Findings of the Association for Computational Linguistics: {EMNLP} 2020, Online Event, 16-20 November 2020}, volume {EMNLP} 2020 of \emph{Findings of {ACL}}, pages 1823--1840. Association for Computational Linguistics.

\bibitem[{Luo and Specia(2024)}]{luo2024understanding}
Haoyan Luo and Lucia Specia. 2024.
\newblock \href {https://doi.org/10.48550/ARXIV.2401.12874} {From understanding to utilization: {A} survey on explainability for large language models}.
\newblock \emph{CoRR}, abs/2401.12874.

\bibitem[{Narayan et~al.(2018)Narayan, Cohen, and Lapata}]{narayan2018don}
Shashi Narayan, Shay~B. Cohen, and Mirella Lapata. 2018.
\newblock \href {https://doi.org/10.18653/V1/D18-1206} {Don't give me the details, just the summary! topic-aware convolutional neural networks for extreme summarization}.
\newblock In \emph{Proceedings of the 2018 Conference on Empirical Methods in Natural Language Processing, Brussels, Belgium, October 31 - November 4, 2018}, pages 1797--1807. Association for Computational Linguistics.

\bibitem[{OpenAI(2023)}]{achiam2023gpt}
OpenAI. 2023.
\newblock \href {https://doi.org/10.48550/ARXIV.2303.08774} {{GPT-4} technical report}.
\newblock \emph{CoRR}, abs/2303.08774.

\bibitem[{Radford et~al.(2017)Radford, J{\'{o}}zefowicz, and Sutskever}]{radford2017learning}
Alec Radford, Rafal J{\'{o}}zefowicz, and Ilya Sutskever. 2017.
\newblock \href {https://arxiv.org/abs/1704.01444} {Learning to generate reviews and discovering sentiment}.
\newblock \emph{CoRR}, abs/1704.01444.

\bibitem[{Razdaibiedina et~al.(2023)Razdaibiedina, Mao, Hou, Khabsa, Lewis, and Almahairi}]{razdaibiedina2023progressive}
Anastasia Razdaibiedina, Yuning Mao, Rui Hou, Madian Khabsa, Mike Lewis, and Amjad Almahairi. 2023.
\newblock \href {https://openreview.net/forum?id=UJTgQBc91\_} {Progressive prompts: Continual learning for language models}.
\newblock In \emph{The Eleventh International Conference on Learning Representations, {ICLR} 2023, Kigali, Rwanda, May 1-5, 2023}. OpenReview.net.

\bibitem[{Rish(2021)}]{rish2021}
Irina Rish. 2021.
\newblock Continual learning with deep architectures.

\bibitem[{Sanh et~al.(2022)Sanh, Webson, Raffel, Bach, Sutawika, Alyafeai, Chaffin, Stiegler, Raja, Dey, Bari, Xu, Thakker, Sharma, Szczechla, Kim, Chhablani, Nayak, Datta, Chang, Jiang, Wang, Manica, Shen, Yong, Pandey, Bawden, Wang, Neeraj, Rozen, Sharma, Santilli, F{\'{e}}vry, Fries, Teehan, Scao, Biderman, Gao, Wolf, and Rush}]{sanh2021multitask}
Victor Sanh, Albert Webson, Colin Raffel, Stephen~H. Bach, Lintang Sutawika, Zaid Alyafeai, Antoine Chaffin, Arnaud Stiegler, Arun Raja, Manan Dey, M~Saiful Bari, Canwen Xu, Urmish Thakker, Shanya~Sharma Sharma, Eliza Szczechla, Taewoon Kim, Gunjan Chhablani, Nihal~V. Nayak, Debajyoti Datta, Jonathan Chang, Mike~Tian{-}Jian Jiang, Han Wang, Matteo Manica, Sheng Shen, Zheng~Xin Yong, Harshit Pandey, Rachel Bawden, Thomas Wang, Trishala Neeraj, Jos Rozen, Abheesht Sharma, Andrea Santilli, Thibault F{\'{e}}vry, Jason~Alan Fries, Ryan Teehan, Teven~Le Scao, Stella Biderman, Leo Gao, Thomas Wolf, and Alexander~M. Rush. 2022.
\newblock \href {https://openreview.net/forum?id=9Vrb9D0WI4} {Multitask prompted training enables zero-shot task generalization}.
\newblock In \emph{The Tenth International Conference on Learning Representations, {ICLR} 2022, Virtual Event, April 25-29, 2022}. OpenReview.net.

\bibitem[{Scao et~al.(2022)Scao, Fan, Akiki, Pavlick, Ilic, Hesslow, Castagn{\'{e}}, Luccioni, Yvon, Gall{\'{e}}, Tow, Rush, Biderman, Webson, Ammanamanchi, Wang, Sagot, Muennighoff, del Moral, Ruwase, Bawden, Bekman, McMillan{-}Major, Beltagy, Nguyen, Saulnier, Tan, Suarez, Sanh, Lauren{\c{c}}on, Jernite, Launay, Mitchell, Raffel, Gokaslan, Simhi, Soroa, Aji, Alfassy, Rogers, Nitzav, Xu, Mou, Emezue, Klamm, Leong, van Strien, Adelani, and et~al.}]{le2023bloom}
Teven~Le Scao, Angela Fan, Christopher Akiki, Ellie Pavlick, Suzana Ilic, Daniel Hesslow, Roman Castagn{\'{e}}, Alexandra~Sasha Luccioni, Fran{\c{c}}ois Yvon, Matthias Gall{\'{e}}, Jonathan Tow, Alexander~M. Rush, Stella Biderman, Albert Webson, Pawan~Sasanka Ammanamanchi, Thomas Wang, Beno{\^{\i}}t Sagot, Niklas Muennighoff, Albert~Villanova del Moral, Olatunji Ruwase, Rachel Bawden, Stas Bekman, Angelina McMillan{-}Major, Iz~Beltagy, Huu Nguyen, Lucile Saulnier, Samson Tan, Pedro~Ortiz Suarez, Victor Sanh, Hugo Lauren{\c{c}}on, Yacine Jernite, Julien Launay, Margaret Mitchell, Colin Raffel, Aaron Gokaslan, Adi Simhi, Aitor Soroa, Alham~Fikri Aji, Amit Alfassy, Anna Rogers, Ariel~Kreisberg Nitzav, Canwen Xu, Chenghao Mou, Chris Emezue, Christopher Klamm, Colin Leong, Daniel van Strien, David~Ifeoluwa Adelani, and et~al. 2022.
\newblock \href {https://doi.org/10.48550/ARXIV.2211.05100} {{BLOOM:} {A} 176b-parameter open-access multilingual language model}.
\newblock \emph{CoRR}, abs/2211.05100.

\bibitem[{Scialom et~al.(2022)Scialom, Chakrabarty, and Muresan}]{scialom2022fine}
Thomas Scialom, Tuhin Chakrabarty, and Smaranda Muresan. 2022.
\newblock \href {https://doi.org/10.18653/V1/2022.EMNLP-MAIN.410} {Fine-tuned language models are continual learners}.
\newblock In \emph{Proceedings of the 2022 Conference on Empirical Methods in Natural Language Processing, {EMNLP} 2022, Abu Dhabi, United Arab Emirates, December 7-11, 2022}, pages 6107--6122. Association for Computational Linguistics.

\bibitem[{Shen et~al.(2023)Shen, Jin, Huang, Liu, Dong, Guo, Wu, Liu, and Xiong}]{DBLP:journals/corr/abs-2309-15025}
Tianhao Shen, Renren Jin, Yufei Huang, Chuang Liu, Weilong Dong, Zishan Guo, Xinwei Wu, Yan Liu, and Deyi Xiong. 2023.
\newblock \href {https://doi.org/10.48550/ARXIV.2309.15025} {Large language model alignment: {A} survey}.
\newblock \emph{CoRR}, abs/2309.15025.

\bibitem[{Shi et~al.(2024)Shi, Jin, Shen, Dong, Wu, and Xiong}]{DBLP:journals/corr/abs-2406-18406}
Dan Shi, Renren Jin, Tianhao Shen, Weilong Dong, Xinwei Wu, and Deyi Xiong. 2024.
\newblock \href {https://doi.org/10.48550/ARXIV.2406.18406} {{IRCAN:} mitigating knowledge conflicts in {LLM} generation via identifying and reweighting context-aware neurons}.
\newblock \emph{CoRR}, abs/2406.18406.

\bibitem[{Simonyan et~al.(2014)Simonyan, Vedaldi, and Zisserman}]{simonyan2013deep}
Karen Simonyan, Andrea Vedaldi, and Andrew Zisserman. 2014.
\newblock \href {http://arxiv.org/abs/1312.6034} {Deep inside convolutional networks: Visualising image classification models and saliency maps}.
\newblock In \emph{2nd International Conference on Learning Representations, {ICLR} 2014, Banff, AB, Canada, April 14-16, 2014, Workshop Track Proceedings}.

\bibitem[{Socher et~al.(2013)Socher, Perelygin, Wu, Chuang, Manning, Ng, and Potts}]{socher2013recursive}
Richard Socher, Alex Perelygin, Jean Wu, Jason Chuang, Christopher~D. Manning, Andrew~Y. Ng, and Christopher Potts. 2013.
\newblock \href {https://aclanthology.org/D13-1170/} {Recursive deep models for semantic compositionality over a sentiment treebank}.
\newblock In \emph{Proceedings of the 2013 Conference on Empirical Methods in Natural Language Processing, {EMNLP} 2013, 18-21 October 2013, Grand Hyatt Seattle, Seattle, Washington, USA, {A} meeting of SIGDAT, a Special Interest Group of the {ACL}}, pages 1631--1642. {ACL}.

\bibitem[{Su et~al.(2020)Su, Guo, Tan, and Chen}]{su2019generative}
Xin Su, Shangqi Guo, Tian Tan, and Feng Chen. 2020.
\newblock \href {https://doi.org/10.1109/TNNLS.2019.2927369} {Generative memory for lifelong learning}.
\newblock \emph{{IEEE} Trans. Neural Networks Learn. Syst.}, 31(6):1884--1898.

\bibitem[{Tang et~al.(2024)Tang, Luo, Huang, Zhang, Wang, Zhao, Wei, and Wen}]{tang2024language}
Tianyi Tang, Wenyang Luo, Haoyang Huang, Dongdong Zhang, Xiaolei Wang, Xin Zhao, Furu Wei, and Ji{-}Rong Wen. 2024.
\newblock \href {https://doi.org/10.18653/V1/2024.ACL-LONG.309} {Language-specific neurons: The key to multilingual capabilities in large language models}.
\newblock In \emph{Proceedings of the 62nd Annual Meeting of the Association for Computational Linguistics (Volume 1: Long Papers), {ACL} 2024, Bangkok, Thailand, August 11-16, 2024}, pages 5701--5715. Association for Computational Linguistics.

\bibitem[{Touvron et~al.(2023)Touvron, Martin, Stone, Albert, Almahairi, Babaei, Bashlykov, Batra, Bhargava, Bhosale, Bikel, Blecher, Canton{-}Ferrer, Chen, Cucurull, Esiobu, Fernandes, Fu, Fu, Fuller, Gao, Goswami, Goyal, Hartshorn, Hosseini, Hou, Inan, Kardas, Kerkez, Khabsa, Kloumann, Korenev, Koura, Lachaux, Lavril, Lee, Liskovich, Lu, Mao, Martinet, Mihaylov, Mishra, Molybog, Nie, Poulton, Reizenstein, Rungta, Saladi, Schelten, Silva, Smith, Subramanian, Tan, Tang, Taylor, Williams, Kuan, Xu, Yan, Zarov, Zhang, Fan, Kambadur, Narang, Rodriguez, Stojnic, Edunov, and Scialom}]{touvron2023llama}
Hugo Touvron, Louis Martin, Kevin Stone, Peter Albert, Amjad Almahairi, Yasmine Babaei, Nikolay Bashlykov, Soumya Batra, Prajjwal Bhargava, Shruti Bhosale, Dan Bikel, Lukas Blecher, Cristian Canton{-}Ferrer, Moya Chen, Guillem Cucurull, David Esiobu, Jude Fernandes, Jeremy Fu, Wenyin Fu, Brian Fuller, Cynthia Gao, Vedanuj Goswami, Naman Goyal, Anthony Hartshorn, Saghar Hosseini, Rui Hou, Hakan Inan, Marcin Kardas, Viktor Kerkez, Madian Khabsa, Isabel Kloumann, Artem Korenev, Punit~Singh Koura, Marie{-}Anne Lachaux, Thibaut Lavril, Jenya Lee, Diana Liskovich, Yinghai Lu, Yuning Mao, Xavier Martinet, Todor Mihaylov, Pushkar Mishra, Igor Molybog, Yixin Nie, Andrew Poulton, Jeremy Reizenstein, Rashi Rungta, Kalyan Saladi, Alan Schelten, Ruan Silva, Eric~Michael Smith, Ranjan Subramanian, Xiaoqing~Ellen Tan, Binh Tang, Ross Taylor, Adina Williams, Jian~Xiang Kuan, Puxin Xu, Zheng Yan, Iliyan Zarov, Yuchen Zhang, Angela Fan, Melanie Kambadur, Sharan Narang, Aur{\'{e}}lien Rodriguez, Robert Stojnic, Sergey Edunov,
  and Thomas Scialom. 2023.
\newblock \href {https://doi.org/10.48550/ARXIV.2307.09288} {Llama 2: Open foundation and fine-tuned chat models}.
\newblock \emph{CoRR}, abs/2307.09288.

\bibitem[{von Oswald et~al.(2023)von Oswald, Niklasson, Randazzo, Sacramento, Mordvintsev, Zhmoginov, and Vladymyrov}]{von2023transformers}
Johannes von Oswald, Eyvind Niklasson, Ettore Randazzo, Jo{\~{a}}o Sacramento, Alexander Mordvintsev, Andrey Zhmoginov, and Max Vladymyrov. 2023.
\newblock \href {https://proceedings.mlr.press/v202/von-oswald23a.html} {Transformers learn in-context by gradient descent}.
\newblock In \emph{International Conference on Machine Learning, {ICML} 2023, 23-29 July 2023, Honolulu, Hawaii, {USA}}, volume 202 of \emph{Proceedings of Machine Learning Research}, pages 35151--35174. {PMLR}.

\bibitem[{Wang et~al.(2019)Wang, Singh, Michael, Hill, Levy, and Bowman}]{wang2018glue}
Alex Wang, Amanpreet Singh, Julian Michael, Felix Hill, Omer Levy, and Samuel~R. Bowman. 2019.
\newblock \href {https://openreview.net/forum?id=rJ4km2R5t7} {{GLUE:} {A} multi-task benchmark and analysis platform for natural language understanding}.
\newblock In \emph{7th International Conference on Learning Representations, {ICLR} 2019, New Orleans, LA, USA, May 6-9, 2019}. OpenReview.net.

\bibitem[{Wang et~al.(2024)Wang, Zhang, Su, and Zhu}]{wang2024comprehensive}
Liyuan Wang, Xingxing Zhang, Hang Su, and Jun Zhu. 2024.
\newblock \href {https://doi.org/10.1109/TPAMI.2024.3367329} {A comprehensive survey of continual learning: Theory, method and application}.
\newblock \emph{{IEEE} Trans. Pattern Anal. Mach. Intell.}, 46(8):5362--5383.

\bibitem[{Wang et~al.(2023{\natexlab{a}})Wang, Chen, Ge, Xia, Bao, Zheng, Zhang, Gui, and Huang}]{wang2023orthogonal}
Xiao Wang, Tianze Chen, Qiming Ge, Han Xia, Rong Bao, Rui Zheng, Qi~Zhang, Tao Gui, and Xuanjing Huang. 2023{\natexlab{a}}.
\newblock \href {https://doi.org/10.18653/V1/2023.FINDINGS-EMNLP.715} {Orthogonal subspace learning for language model continual learning}.
\newblock In \emph{Findings of the Association for Computational Linguistics: {EMNLP} 2023, Singapore, December 6-10, 2023}, pages 10658--10671. Association for Computational Linguistics.

\bibitem[{Wang et~al.(2022)Wang, Wen, Zhang, Hou, Liu, and Li}]{wang2022finding}
Xiaozhi Wang, Kaiyue Wen, Zhengyan Zhang, Lei Hou, Zhiyuan Liu, and Juanzi Li. 2022.
\newblock \href {https://doi.org/10.18653/V1/2022.EMNLP-MAIN.765} {Finding skill neurons in pre-trained transformer-based language models}.
\newblock In \emph{Proceedings of the 2022 Conference on Empirical Methods in Natural Language Processing, {EMNLP} 2022, Abu Dhabi, United Arab Emirates, December 7-11, 2022}, pages 11132--11152. Association for Computational Linguistics.

\bibitem[{Wang et~al.(2023{\natexlab{b}})Wang, Liu, Ji, Wang, Wu, Jiang, Chao, Han, Wang, Shao, and Zeng}]{wang2023rehearsal}
Zhicheng Wang, Yufang Liu, Tao Ji, Xiaoling Wang, Yuanbin Wu, Congcong Jiang, Ye~Chao, Zhencong Han, Ling Wang, Xu~Shao, and Wenqiu Zeng. 2023{\natexlab{b}}.
\newblock \href {https://doi.org/10.18653/V1/2023.ACL-LONG.612} {Rehearsal-free continual language learning via efficient parameter isolation}.
\newblock In \emph{Proceedings of the 61st Annual Meeting of the Association for Computational Linguistics (Volume 1: Long Papers), {ACL} 2023, Toronto, Canada, July 9-14, 2023}, pages 10933--10946. Association for Computational Linguistics.

\bibitem[{Wei et~al.(2022)Wei, Bosma, Zhao, Guu, Yu, Lester, Du, Dai, and Le}]{wei2021finetuned}
Jason Wei, Maarten Bosma, Vincent~Y. Zhao, Kelvin Guu, Adams~Wei Yu, Brian Lester, Nan Du, Andrew~M. Dai, and Quoc~V. Le. 2022.
\newblock \href {https://openreview.net/forum?id=gEZrGCozdqR} {Finetuned language models are zero-shot learners}.
\newblock In \emph{The Tenth International Conference on Learning Representations, {ICLR} 2022, Virtual Event, April 25-29, 2022}. OpenReview.net.

\bibitem[{Welbl et~al.(2017)Welbl, Liu, and Gardner}]{welbl2017crowdsourcing}
Johannes Welbl, Nelson~F. Liu, and Matt Gardner. 2017.
\newblock \href {https://doi.org/10.18653/V1/W17-4413} {Crowdsourcing multiple choice science questions}.
\newblock In \emph{Proceedings of the 3rd Workshop on Noisy User-generated Text, NUT@EMNLP 2017, Copenhagen, Denmark, September 7, 2017}, pages 94--106. Association for Computational Linguistics.

\bibitem[{Williams et~al.(2018)Williams, Nangia, and Bowman}]{williams2017broad}
Adina Williams, Nikita Nangia, and Samuel~R. Bowman. 2018.
\newblock \href {https://doi.org/10.18653/V1/N18-1101} {A broad-coverage challenge corpus for sentence understanding through inference}.
\newblock In \emph{Proceedings of the 2018 Conference of the North American Chapter of the Association for Computational Linguistics: Human Language Technologies, {NAACL-HLT} 2018, New Orleans, Louisiana, USA, June 1-6, 2018, Volume 1 (Long Papers)}, pages 1112--1122. Association for Computational Linguistics.

\bibitem[{Wu et~al.(2024)Wu, Dong, Xu, and Xiong}]{DBLP:conf/acl/WuDXX24}
Xinwei Wu, Weilong Dong, Shaoyang Xu, and Deyi Xiong. 2024.
\newblock \href {https://doi.org/10.18653/V1/2024.FINDINGS-ACL.315} {Mitigating privacy seesaw in large language models: Augmented privacy neuron editing via activation patching}.
\newblock In \emph{Findings of the Association for Computational Linguistics, {ACL} 2024, Bangkok, Thailand and virtual meeting, August 11-16, 2024}, pages 5319--5332. Association for Computational Linguistics.

\bibitem[{Wu et~al.(2023{\natexlab{a}})Wu, Li, Xu, Dong, Wu, Bian, and Xiong}]{wu2023depn}
Xinwei Wu, Junzhuo Li, Minghui Xu, Weilong Dong, Shuangzhi Wu, Chao Bian, and Deyi Xiong. 2023{\natexlab{a}}.
\newblock \href {https://doi.org/10.18653/V1/2023.EMNLP-MAIN.174} {{DEPN:} detecting and editing privacy neurons in pretrained language models}.
\newblock In \emph{Proceedings of the 2023 Conference on Empirical Methods in Natural Language Processing, {EMNLP} 2023, Singapore, December 6-10, 2023}, pages 2875--2886. Association for Computational Linguistics.

\bibitem[{Wu et~al.(2023{\natexlab{b}})Wu, Zhao, Li, Qin, and Xiong}]{wu2023improving}
Yang Wu, Yanyan Zhao, Zhongyang Li, Bing Qin, and Kai Xiong. 2023{\natexlab{b}}.
\newblock \href {https://doi.org/10.48550/ARXIV.2305.04429} {Improving cross-task generalization with step-by-step instructions}.
\newblock \emph{CoRR}, abs/2305.04429.

\bibitem[{Xie et~al.(2021)Xie, Feng, Gu, and Yu}]{xie2021importance}
Wanying Xie, Yang Feng, Shuhao Gu, and Dong Yu. 2021.
\newblock \href {https://doi.org/10.18653/V1/2021.ACL-LONG.445} {Importance-based neuron allocation for multilingual neural machine translation}.
\newblock In \emph{Proceedings of the 59th Annual Meeting of the Association for Computational Linguistics and the 11th International Joint Conference on Natural Language Processing, {ACL/IJCNLP} 2021, (Volume 1: Long Papers), Virtual Event, August 1-6, 2021}, pages 5725--5737. Association for Computational Linguistics.

\bibitem[{Xiong et~al.(2019)Xiong, Wu, Wang, Kulkarni, Yu, Chang, Guo, and Wang}]{xiong2019tweetqa}
Wenhan Xiong, Jiawei Wu, Hong Wang, Vivek Kulkarni, Mo~Yu, Shiyu Chang, Xiaoxiao Guo, and William~Yang Wang. 2019.
\newblock \href {https://doi.org/10.18653/V1/P19-1496} {{TWEETQA:} {A} social media focused question answering dataset}.
\newblock In \emph{Proceedings of the 57th Conference of the Association for Computational Linguistics, {ACL} 2019, Florence, Italy, July 28- August 2, 2019, Volume 1: Long Papers}, pages 5020--5031. Association for Computational Linguistics.

\bibitem[{Xu et~al.(2024)Xu, Zhan, Wong, and Chao}]{xu2024let}
Haoyun Xu, Runzhe Zhan, Derek~F. Wong, and Lidia~S. Chao. 2024.
\newblock \href {https://doi.org/10.48550/ARXIV.2403.11621} {Let's focus on neuron: Neuron-level supervised fine-tuning for large language model}.
\newblock \emph{CoRR}, abs/2403.11621.

\bibitem[{Yang et~al.(2024)Yang, Zhang, Xu, Lu, Heng, and Lam}]{yang2024unveiling}
Haoran Yang, Yumeng Zhang, Jiaqi Xu, Hongyuan Lu, Pheng{-}Ann Heng, and Wai Lam. 2024.
\newblock \href {https://doi.org/10.18653/V1/2024.NAACL-LONG.51} {Unveiling the generalization power of fine-tuned large language models}.
\newblock In \emph{Proceedings of the 2024 Conference of the North American Chapter of the Association for Computational Linguistics: Human Language Technologies (Volume 1: Long Papers), {NAACL} 2024, Mexico City, Mexico, June 16-21, 2024}, pages 884--899. Association for Computational Linguistics.

\bibitem[{Zhang et~al.(2019)Zhang, Baldridge, and He}]{zhang2019paws}
Yuan Zhang, Jason Baldridge, and Luheng He. 2019.
\newblock \href {https://doi.org/10.18653/V1/N19-1131} {{PAWS:} paraphrase adversaries from word scrambling}.
\newblock In \emph{Proceedings of the 2019 Conference of the North American Chapter of the Association for Computational Linguistics: Human Language Technologies, {NAACL-HLT} 2019, Minneapolis, MN, USA, June 2-7, 2019, Volume 1 (Long and Short Papers)}, pages 1298--1308. Association for Computational Linguistics.

\bibitem[{Zhao et~al.(2024)Zhao, Zhang, Chen, Kawaguchi, and Bing}]{zhao2024large}
Yiran Zhao, Wenxuan Zhang, Guizhen Chen, Kenji Kawaguchi, and Lidong Bing. 2024.
\newblock \href {https://doi.org/10.48550/ARXIV.2402.18815} {How do large language models handle multilingualism?}
\newblock \emph{CoRR}, abs/2402.18815.

\bibitem[{Zhou et~al.(2023)Zhou, Liu, Xu, Iyer, Sun, Mao, Ma, Efrat, Yu, Yu, Zhang, Ghosh, Lewis, Zettlemoyer, and Levy}]{zhou2024lima}
Chunting Zhou, Pengfei Liu, Puxin Xu, Srinivasan Iyer, Jiao Sun, Yuning Mao, Xuezhe Ma, Avia Efrat, Ping Yu, Lili Yu, Susan Zhang, Gargi Ghosh, Mike Lewis, Luke Zettlemoyer, and Omer Levy. 2023.
\newblock \href {http://papers.nips.cc/paper\_files/paper/2023/hash/ac662d74829e4407ce1d126477f4a03a-Abstract-Conference.html} {{LIMA:} less is more for alignment}.
\newblock In \emph{Advances in Neural Information Processing Systems 36: Annual Conference on Neural Information Processing Systems 2023, NeurIPS 2023, New Orleans, LA, USA, December 10 - 16, 2023}.

\bibitem[{Zhu et~al.(2024)Zhu, Pan, Li, and Xiong}]{zhu2024landermt}
Shaolin Zhu, Leiyu Pan, Bo~Li, and Deyi Xiong. 2024.
\newblock \href {https://doi.org/10.18653/V1/2024.ACL-LONG.656} {Landermt: Dectecting and routing language-aware neurons for selectively finetuning llms to machine translation}.
\newblock In \emph{Proceedings of the 62nd Annual Meeting of the Association for Computational Linguistics (Volume 1: Long Papers), {ACL} 2024, Bangkok, Thailand, August 11-16, 2024}, pages 12135--12148. Association for Computational Linguistics.

\end{thebibliography}

\appendix
\clearpage
\section{Appendix}
\subsection{Taylor Expansion}
\label{sec:appendix A-111}
We follow \citet{xie2021importance} and \citet{zhu2024landermt} to provide the proof of Equation \ref{equ:3}.

We adopt a criterion based on the Taylor Expansion, where we directly approximate the change in loss when removing a particular neuron. Let $\boldsymbol{\omega}^i_j$ be the output of the $j$-th neuron in layer $i$, and $\Omega$ represents the set of other neurons. Assuming the independence of each neuron in the model, the change of loss when removing the $j$-th neuron in layer $i$ can be represented as:
\begin{equation}
\left|\Delta\mathcal{L}(\boldsymbol{\omega}^i_j)\right|=\left| \mathcal{L}(\Omega, \boldsymbol{\omega}^i_j=0)- \mathcal{L}(\Omega, \boldsymbol{\omega}^i_j)  \right|
\label{equ:4}
\end{equation}
where $\mathcal{L}(\Omega, \boldsymbol{\omega}^i_j=0)$ is the loss value if the $j$-th neuron in layer $i$ is pruned and $\mathcal{L}(\Omega, \boldsymbol{\omega}^i_j)$ is the loss if it is not pruned. For the function $\mathcal{L}(\Omega, \boldsymbol{\omega}^i_j)$, its Taylor Expansion at $\boldsymbol{\omega}^i_j=0$ is:
\begin{equation}
\mathcal{L}(\Omega, \boldsymbol{\omega}^i_j) = \mathcal{L}(\Omega, \boldsymbol{\omega}^i_j=0) + \frac{\partial{\mathcal{L}(\Omega, \boldsymbol{\omega}^i_j)}}{\partial{\boldsymbol{\omega}^i_j}} \boldsymbol{\omega}^i_j + R_1({\boldsymbol{\omega}^i_j})
\label{equ:5}
\end{equation}
where $R_1({\boldsymbol{\omega}^i_j})$ can be ignored since the derivatives of the activation function of second order and higher in the model tend to be zero. So the above equation can be reduced to the following form:
\begin{equation}
\mathcal{L}(\Omega, \boldsymbol{\omega}^i_j) \approx \mathcal{L}(\Omega, \boldsymbol{\omega}^i_j=0) + \frac{\partial{\mathcal{L}(\Omega, \boldsymbol{\omega}^i_j)}}{\partial{\boldsymbol{\omega}^i_j}} \boldsymbol{\omega}^i_j
\label{equ:6}
\end{equation}

Therefore $\left|\Delta\mathcal{L}(\boldsymbol{\omega}^i_j)\right|$ can eventually be simplified to the following form:
\begin{equation}
\left|\Delta\mathcal{L}(\boldsymbol{\omega}^i_j)\right| \approx \left|\frac{\partial{\mathcal{L}(\Omega, \boldsymbol{\omega}^i_j)}}{\partial{\boldsymbol{\omega}^i_j}} \boldsymbol{\omega}^i_j \right|
\label{equ:7}
\end{equation}

\subsection{Details of W-NCFT Method}
\label{sec:appendix A-222}

Assuming that the model has been trained on the previous $i$ tasks, when inference is executed on the $j$-th task ($j \leqslant i$), we calculate the similarity between task $j$ and the previous $i$ tasks. The similarity between any two tasks as follows:

\begin{equation}
\mathrm{sim}(x,y)=\frac{\boldsymbol{fea}_x \cdot \boldsymbol{fea}_y}{\left|\left| \boldsymbol{fea}_x\right|\right| \times \left|\left| \boldsymbol{fea}_y\right|\right|}
\label{equ:8}
\end{equation}
where $\boldsymbol{fea}_{task} \in \mathbb{R}^d$ is the task vector. We randomly select $1000$ samples for each task, and use the Llama-2-7b to compute the mean of the features in the last layer for each particular task sample, and finally take the mean of these sample features as a representation of the task vector. 

Then, we get a similarity vector $(\mathrm{sim}_j^1, \cdots, \mathrm{sim}_j^i)$, where $\mathrm{sim}_j^k$ is the similarity between task $j$ and task $k$ ($1 \leqslant k \leqslant i$). Finally, we conduct $\mathrm{Softmax}$ normalization:
\begin{equation}
\boldsymbol{Sim}_j=\mathrm{Softmax}(\mathrm{sim}_j^1, \cdots, \mathrm{sim}_j^i)
\label{equ:9}
\end{equation}

During inference, for the parameter matrix $\boldsymbol{W}$ of the FFN module in a particular layer of the model, we sequentially identify the task-specific neuron parameters (i.e., certain columns of $\boldsymbol{W}$) among the tasks previously trained, ranging from task $1$ to task $i$, and allocate weights to this portion of parameters based on $\boldsymbol{Sim}_j$ as follows:
\begin{equation}
\boldsymbol{W}'=\sum_{k=1}^i\boldsymbol{Sim}_j[k] \times \boldsymbol{W}_{task-k}
\label{equ:10}
\end{equation}
where $\boldsymbol{W}_{task-k}$ is the task-specific neuron parameter for the $k$-th task, the summation notation $\sum$ indicates that combining the individual submatrices by columns, and $\boldsymbol{W}'$ is the final weighted parameter matrix.

Subsequently, inference is conducted. We refer to this approach as \textbf{W}eighted \textbf{N}euron-level \textbf{C}ontinuous \textbf{F}ine-\textbf{T}uning (W-NCFT).

\subsection{Datasets and Metrics for Identifying Neurons Experiments}
\label{sec:appendix B-000}
According to task output forms, we tested two main types of tasks: \textit{classification} and \textit{generation}.
\begin{itemize}
\item[$\bullet$] For classification tasks, we chose three tasks. They are sentiment classification, including AmazonFood \citep{keung2020multilingual}, SST-2 \citep{socher2013recursive}; paraphrase detection, including QQP \citep{wang2018glue}, Paws \citep{zhang2019paws}; and natural language inference, including MNLI \citep{williams2017broad}, GPTNLI\footnote{\url{https://huggingface.co/datasets/pietrolesci/gpt3_nli}}.
\item[$\bullet$] For generation tasks, we chose three tasks. They are summary generation, including CNN/DailyMail \citep{hermann2015teaching}, Xsum \citep{narayan2018don}; question generation, including Sciqa \citep{welbl2017crowdsourcing}, Tweetqa \citep{xiong2019tweetqa}; and data-to-text generation, including E2E \citep{duvsek2020evaluating}, CommonGen \citep{lin2019commongen}.
\end{itemize}

We used accuracy to evaluate classification tasks and Rouge-L\footnote{\url{https://huggingface.co/spaces/evaluate-metric/rouge}} to evaluate generation tasks.

\subsection{Additional Experiments for Identifying Neurons}
\label{sec:appendix B-111}
Table \ref{tab_6} shows the results of the deactivation experiments on Bloom-7b1 and Table \ref{tab_7} shows the results of the fine-tuning experiments on Bloom-7b1. We can find a more significant trend for fine-tuning and deactivation of task-specific neurons compared to randomly selected neurons, consistent with the observation in Llama-2-7b.

\begin{figure*}[t]
  \centering
    \subfigure[Deactivation on classification tasks]{
       \centering
       \includegraphics[width=0.4 \textwidth]{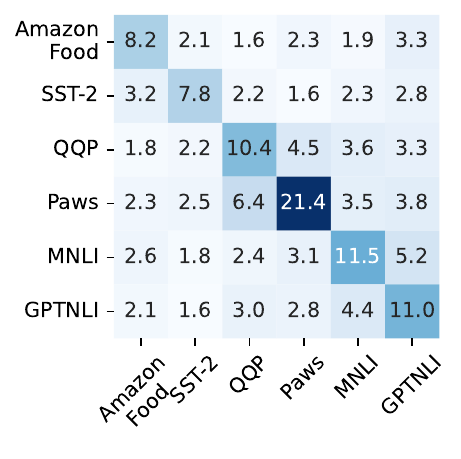}
    }
    \subfigure[Deactivation on generation tasks]{
        \centering
        \includegraphics[width=0.4 \textwidth]{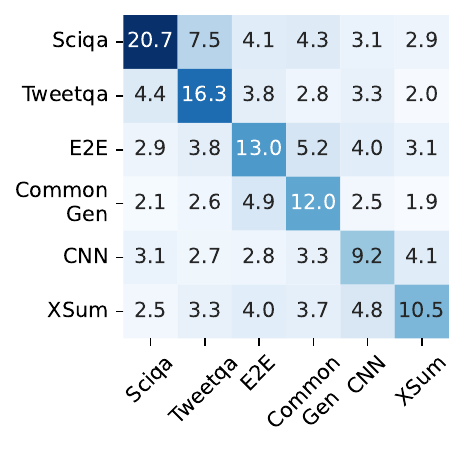}
    }
    \subfigure[Fine-tuning on classification tasks]{
       \centering
       \includegraphics[width=0.4 \textwidth]{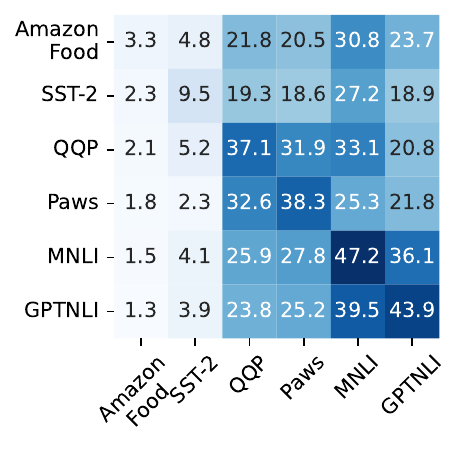}
    }
    \subfigure[Fine-tuning on generation tasks]{
        \centering
        \includegraphics[width=0.4 \textwidth]{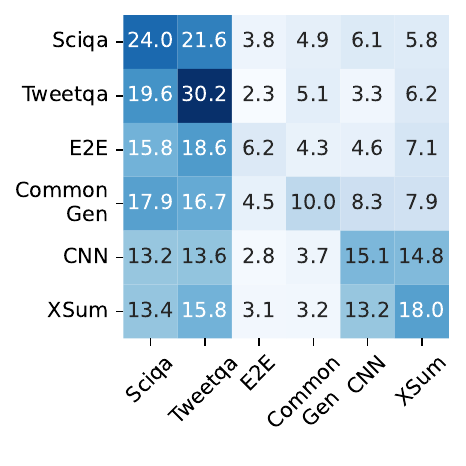}
    }
  \caption{Performance of Llama-2-7b on all tasks after deactivation or fine-tuning a particular class task-specific neurons.   The element in the $i$-th row and $j$-th column is the performance change for task $j$ due to deactivation or fine-tuning of the task $i$ specific neurons.}
  \label{fig5}
\end{figure*}

Figure \ref{fig5} (a) and (b) show the performance on all tasks after deactivating a particular class of task-specific neurons for Llama-2-7b on six classification and six generation tasks, respectively.  In both $6\times6$ matrices, the values on the main diagonal are significantly higher than those at other locations in the same row and column.  This suggests that (1) the impact of deactivating a particular class of task-specific neurons on all other tasks is weaker than the impact on this task itself, (2) the impact of deactivating a particular class of task-specific neurons on this task itself is stronger than the impact of deactivating other classes of task-specific neurons on this task.

Figure \ref{fig5} (c) and (d) show the performance on all tasks after fine-tuning a particular class of task-specific neurons on six classification and six generation tasks for Llama-2-7b, respectively. In both $6\times6$ matrices, most of the values on the main diagonal are also significantly higher than those of the other elements in the same row and column. The few exceptions are the AmazonFood and SST-2 tasks, which are relatively simple and where zero-shot learning works well enough so that there is little space for improvement. There's also E2E and CommonGen, which are limited by the difficulty of tasks and have limited scope for improvement. But in each column of these matrices, the impact of deactivating a particular class of task-specific neurons on this task itself is stronger than the impact of deactivating other classes of task-specific neurons on this task. 

These results are sufficient to show that our experiments eliminate noise among task-specific neurons and ensure the task-specificity of the neurons we identify.

\subsection{Ablation Experiments for Single-task Training}
\label{sec:appendix B-222}
\begin{figure*}[t]
\centering
\includegraphics[width=0.9\linewidth]{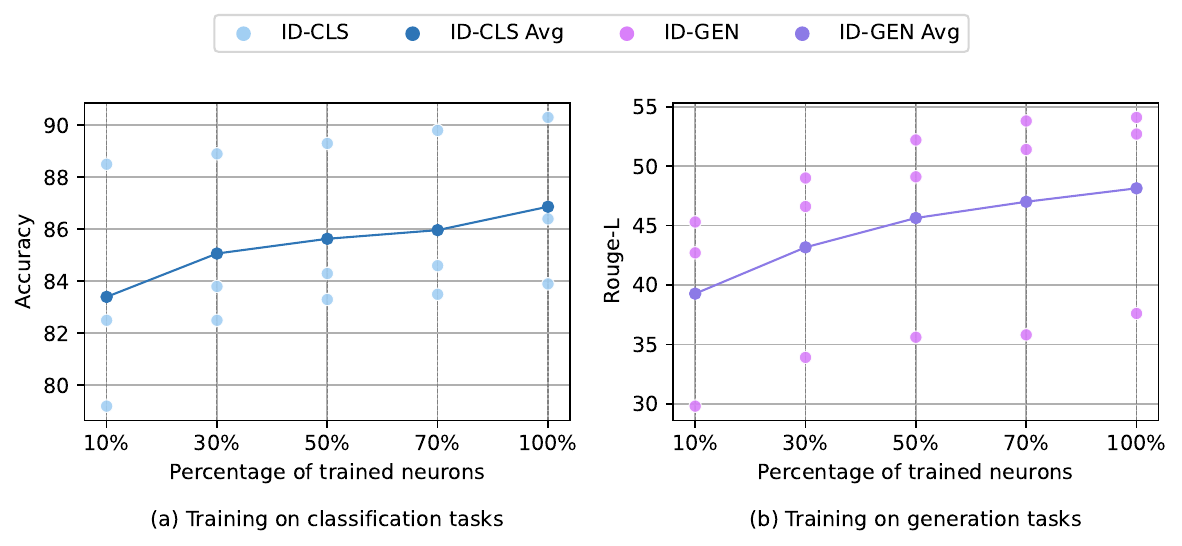}
\caption{Results of training and testing each task individually for observing specialization.}
\label{fig6}
\end{figure*}

Figure \ref{fig6} shows the results of training and testing each task individually.

\subsection{Overlap Rate}
\label{sec:appendix B-333}
We calculate the overlap rate of task-specific neurons between the training tasks and test tasks as:
\begin{equation}
\mathrm{overlap}(x, y)=\frac{\mathcal{N}_x \cap \mathcal{N}_y}{\mathcal{N}_x \cup \mathcal{N}_y}
\label{equ:11}
\end{equation}
where $\mathcal{N}_{tasks}$ denotes the set of task-specific neurons.

Table \ref{tab_8} shows the overlap rate of task-specific neurons between the training tasks and test tasks. It is worth noting that for all training-test task combinations, we use the overall set of task-specific neurons of three training tasks as $\mathcal{N}_x$ and the overall set of task-specific neurons of three test tasks as $\mathcal{N}_y$.

\subsection{Ablation Experiments on Overlap Rates and Fine-tuning Proportions}
\label{sec:appendix B-444}

Specifically, we chose the fine-tuning neuron proportions as $10\%$, $30\%$, and $50\%$. Under each proportion, we set multiple overlap rates of trained neurons and test task neurons. For example, at a fine-tuning neuron proportion of $10\%$, we first calculate the total number of neurons that need to be trained at this time. We then divide the trained neurons into two sets, a set of task-specific neurons for the test task, and another set containing all remaining neurons. According to the preset overlap rate and the total number of trained neurons, we are able to calculate the number of neurons to be selected from these two sets, and we randomly select them in each of the two sets.

Tables \ref{tab_9}, \ref{tab_10} and \ref{tab_11} show the results of the three sets of experiments for classification - classification, generation - generation, and generation - classification, respectively. It can be found that when the proportion of trained neurons is fixed, the performance is improving as the overlap rate increases, which directly proves the conclusion of our paper. In addition to this, when the overlap rate is fixed, the performance is improving as the total number of trained neurons increases. This can be interpreted as a gain from an increase in the number of trained parameters. It is worth noting that when the overlap rate is not fixed, the performance may not be as good as training a small number of neurons despite training more neurons. For example, in the classification - classification experiment, with $30\%$ of trained neurons and an overlap rate of $10\%$, the performance is $81.4$. However, with $10\%$ of trained neurons and an overlap rate of $70\%$, the performance is $82.0$. In all three sets of experiments, the above conclusions hold.

\subsection{Benchmarks of Continuous Learning}
\label{sec:appendix C-111}
Table \ref{tab_12} and Table \ref{tab_13} show the datasets included in the Standard CL Benchmark and Large Number of Tasks Benchmark, respectively. Note that the original Large Number of Tasks Benchmark have $15$ tasks, from which we select $8$ tasks to form a simplified version for our experiments.

Table \ref{tab_14} shows the task order sequence for the two continuous learning benchmarks.

\begin{table*}[]
\centering
\resizebox{0.9\textwidth}{!}{
\begin{tabular}{l|cccccc|c}
   \toprule
   \textbf{Method \textbackslash \ Task-CLS} & \textbf{AmazonFood} & \textbf{SST-2} & \textbf{QQP} &  \textbf{Paws} & \textbf{MNLI} & \textbf{GPTNLI} & \textbf{Avg.}\\
   \midrule
   Original & 90.6 & 91.2 & 81.8 & 91 & 80.3 & 79.5 & 85.7 \\ 
   Deactivate-Random & 89.5 & 89.7 & 79.3 & 88.5 & 78.5 & 77.6 & 83.9 \\ 
   Deactivate-Task & \textbf{80.3} & \textbf{83.5} & \textbf{71.2} & \textbf{82.3} & \textbf{70.6} & \textbf{69.5} & \textbf{76.2} \\ 
   \toprule
   \textbf{Method \textbackslash \ Task-GEN} & \textbf{Sciqa} & \textbf{Tweetqa} & \textbf{E2E} &  \textbf{CommonGen} & \textbf{CNN/DailyMail} & \textbf{XSum} & \textbf{Avg.}\\
   \midrule
   Original & 53.8 & 41.8 & 54.5 & 45.6 & 31.8 & 33.2 & 43.5 \\ 
   Deactivate-Random & 50.9 & 40.8 & 52.5 & 41.6 & 29.8 & 30.8 & 41.1 \\ 
   Deactivate-Task & \textbf{34.7} & \textbf{30.6} & \textbf{41.8} & \textbf{32.3} & \textbf{20.7} & \textbf{21.5} & \textbf{30.3} \\
   \bottomrule
\end{tabular}}
\caption{Performance of Bloom-7b1 after task-specific neurons deactivation or without deactivation in each task. \enquote{Original} is the performance after fine-tuning with multi-task data without any neurons being deactivated. \enquote{Deactivate-Task} indicates deactivation of task-specific neurons. \enquote{Deactivate-Random} indicates that the same number of neurons are randomly selected for deactivation. Task-CLS: Classification Task. Task-GEN: Generation Task.}
\label{tab_6}
\end{table*}

\begin{table*}[h]
\centering
\resizebox{0.9\textwidth}{!}{
\begin{tabular}{l|cccccc|c}
   \toprule
   \textbf{Method \textbackslash \ Task-CLS} & \textbf{AmazonFood} & \textbf{SST-2} & \textbf{QQP} &  \textbf{Paws} & \textbf{MNLI} & \textbf{GPTNLI} & \textbf{Avg.}\\
   \midrule
   Zero-shot & 83.7 & 79.1 & 46.5 & 44.3 & 33.6 & 34.2 & 53.6 \\
   Train-Random & 84.1 & 80.5 & 48.0 & 46.1 & 35.2 & 36.1 & 55.0 \\
   Train-Task & \textbf{87.6} & \textbf{88.3} & \textbf{77.6} & \textbf{82.3} & \textbf{79.4} & \textbf{72.0} & \textbf{81.2}  \\
   \toprule
   \textbf{Method \textbackslash \ Task-GEN} & \textbf{Sciqa} & \textbf{Tweetqa} & \textbf{E2E} &  \textbf{CommonGen} & \textbf{CNN/DailyMail} & \textbf{XSum} & \textbf{Avg.}\\
   \midrule
   Zero-shot & 23.1 & 10.3 & 33.2 & 23.6 & 12.5 & 13.4 & 19.4   \\
   Train-Random & 23.8 & 12.7 & 34.8 & 25.2 & 14.2 & 15.5 & 21.0  \\
   Train-Task & \textbf{42.0} & \textbf{34.3} & \textbf{40.4} & \textbf{33.0} & \textbf{27.1} & \textbf{28.6} & \textbf{34.2}  \\
   \bottomrule
\end{tabular}}
\caption{Performance of Bloom-7b1 after fine-tuning task-specific neurons and under the zero-shot setting. \enquote{Train-Task} indicates training task-specific neurons. \enquote{Train-Random} indicates that the same number of neurons are randomly selected for training. Task-CLS: Classification Task. Task-GEN: Generation Task.}
\label{tab_7}
\end{table*}

\clearpage

\begin{table}[t]
\centering
\resizebox{0.5\textwidth}{!}{
\begin{tabular}{l|ccccc}
   \toprule
   \textbf{Group} & \textbf{10\%} & \textbf{30\%} & \textbf{50\%} &  \textbf{70\%} & \textbf{100\%}\\
   \midrule
   CLS-CLS & 20.8 & 53.9 & 84.5 & 96.2 & 100 \\
   CLS-GEN & 12.9 & 41.6 & 71.5 & 83.5 & 100 \\
   GEN-CLS & 11.8 & 40.2 & 69.3 & 81.8 & 100 \\
   GEN-GEN & 21.6 & 52.5 & 82.0 & 94.3 & 100 \\
   \bottomrule
\end{tabular}}
\caption{The overlap rate of task-specific neurons between training tasks and test tasks when controlling the proportion of task-specific neurons.}
\label{tab_8}
\end{table}

\begin{table}[]
\centering
\resizebox{0.5\textwidth}{!}{
\begin{tabular}{l|ccc}
   \toprule
   \textbf{Overlap rate \textbackslash \ Percentage of trained neurons} & \textbf{10\%} & \textbf{30\%} & \textbf{50\%}\\
   \midrule
   10\% & 80.2 & 81.4 & 81.8 \\ 
   \textit{20.8\%} & \textit{80.7} & - & - \\ 
   30\% & 81.1 & 82.0 & 82.3 \\
   50\% & 81.5 & 82.3 & 82.8 \\
   \textit{53.9\%} & - & \textit{82.5} & - \\
   70\% & 82.0 & 82.7 & 83.0 \\
   \textit{84.5\%} & - & - & \textit{83.1} \\
   100\% & 82.2 & 83.1 & 83.6 \\
   \bottomrule
\end{tabular}}
\caption{Results at different fine-tuned neuron proportions ($10\%$, $30\%$, $50\%$) controlling the overlap rate under the classification - classification combination. Italics indicate the original experimental results and overlap rates.}
\label{tab_9}
\end{table}

\begin{table}[]
\centering
\resizebox{0.5\textwidth}{!}{
\begin{tabular}{l|ccc}
   \toprule
   \textbf{Overlap rate \textbackslash \ Percentage of trained neurons} & \textbf{10\%} & \textbf{30\%} & \textbf{50\%}\\
   \midrule
    10\% & 31.6 & 32.1 & 32.3 \\ 
    \textit{21.6\%} & \textit{32.2} & - & - \\
    30\% & 32.5 & 32.9 & 33.5 \\
    50\% & 32.7 & 33.4 & 33.8 \\ 
    \textit{52.5\%} & - & \textit{33.8} & - \\
    70\% & 32.9 & 34.0 & 34.1 \\
    \textit{82.0\%} & - & - & \textit{34.9} \\
    100\% & 33.1 & 34.4 & 35.1 \\
   \bottomrule
\end{tabular}}
\caption{Results at different fine-tuned neuron proportions ($10\%$, $30\%$, $50\%$) controlling the overlap rate under the generation - generation combination. Italics indicate the original experimental results and overlap rates.}
\label{tab_10}
\end{table}

\begin{table}[]
\centering
\resizebox{0.5\textwidth}{!}{
\begin{tabular}{l|ccc}
   \toprule
   \textbf{Overlap rate \textbackslash \ Percentage of trained neurons} & \textbf{10\%} & \textbf{30\%} & \textbf{50\%}\\
   \midrule
    10\% & 78.2 & 78.5 & 78.6 \\ 
    \textit{11.8\%} & \textit{78.4} & - & - \\ 
    30\% & 78.8 & 78.7 & 79.3 \\ 
    \textit{40.2\%} & - & \textit{78.7} & - \\
    50\% & 79.0 & 79.4 & 79.7 \\ 
    \textit{69.3\%} & - & - & \textit{80.1} \\
    70\% & 79.3 & 79.6 & 80.3 \\
    100\% & 79.5 & 79.9 & 80.8 \\
   \bottomrule
\end{tabular}}
\caption{Results at different fine-tuned neuron proportions ($10\%$, $30\%$, $50\%$) controlling the overlap rate under the generation - classification combination. Italics indicate the original experimental results and overlap rates.}
\label{tab_11}
\end{table}

\begin{table}[h]
\centering
\resizebox{0.5\textwidth}{!}{
\begin{tabular}{lccc}
   \toprule
   \textbf{Dataset} & \textbf{Class} & \textbf{Task Type} & \textbf{Domain} \\
   \midrule
   AGNews & 4 & Topic classification & News\\
   Amazon & 5 & Sentiment anlysis & Amazon reviews \\
   DBPedia & 14 & Topic classification &  Wikipedia \\
   Yahoo & 10 & Q\&A & Yahoo Q\&A \\
   \bottomrule
\end{tabular}}
\caption{Details of the Standard CL Benchmark.}
\label{tab_12}
\end{table}

\begin{table}[h]
\centering
\resizebox{0.5\textwidth}{!}{
\begin{tabular}{lccc}
   \toprule
   \textbf{Dataset} & \textbf{Class} & \textbf{Task Type} & \textbf{Domain} \\
   \midrule
   Amazon & 5 & Sentiment anlysis & Amazon reviews \\
   DBPedia & 14 & Topic classification &  Wikipedia \\
   Yahoo & 10 & Q\&A & Yahoo Q\&A \\
   AGNews & 4 & Topic classification & News\\
   MNLI & 3 & NLI  & various \\
   QQP & 2 & Paragraph detection & Quora \\
   RTE & 2 & NLI & news, Wikipedia  \\
   SST-2 & 2 & Sentiment analysis & movie reviews \\
   \bottomrule
\end{tabular}}
\caption{Details of the simplified version Large Number of Tasks Benchmark.}
\label{tab_13}
\end{table}

\begin{table}[h]
\centering
\resizebox{0.5\textwidth}{!}{
\begin{tabular}{cc}
   \toprule
   \textbf{Order} & \textbf{Task Sequence}  \\
   \midrule
   1 &  DBPedia → Amazon → Yahoo → AGNews \\
   2 &  DBPedia → Amazon → AGNews → Yahoo \\
   3 &  Yahoo → Amazon → AGNews → DBPedia \\
   \midrule
   4 &  \begin{tabular}[c]{@{}l@{}}MNLI → QQP → RTE → Amazon → \\
   SST-2 → DBPedia → AGNews → Yahoo\end{tabular}\\
   5 &  \begin{tabular}[c]{@{}l@{}}Amazon → AGNews → Yahoo → QQP → \\
   RTE → MNLI → DBPedia → SST-2\end{tabular}\\
   6 &  \begin{tabular}[c]{@{}l@{}}AGNews → Yahoo → SST-2 → RTE → \\
   QQP → MNLI → DBPedia → Amazon\end{tabular}\\
   \bottomrule
\end{tabular}}
\caption{Task order sequence for two continuous learning benchmarks.}
\label{tab_14}
\end{table}

\end{document}